\documentclass[10pt, a4paper]{article}

\usepackage{graphicx}
\usepackage{commath}
\usepackage{amsthm}
\usepackage{hyperref}
\usepackage{color}
\usepackage[table]{xcolor}
\usepackage{multirow}
\usepackage{booktabs}
\usepackage{subfigure}
\usepackage{float}

\usepackage[english]{babel}
\usepackage{amssymb,amsmath,amsthm, amsfonts}

\RequirePackage[left=2cm,%
                right=2cm,%
                top=2.25cm,%
                bottom=2.25cm,%
                headheight=12pt,%
                letterpaper]{geometry}%

\title{Dynamic texture analysis with diffusion in networks}

\author{Lucas C. Ribas$^{1,3}$, Wesley N. Gon\c{c}alves$^{2}$, Odemir M. Bruno$^{3,1}$}

\date{}
\begin{document}
\maketitle
\noindent{$^1$Institute of Mathematics and Computer Science, University of S\~ao Paulo, USP, Avenida Trabalhador s\~ao-carlense, 400, 13566-590, S\~ao Carlos, SP, Brazil}

\noindent{$^2$Federal University of Mato Grosso do Sul, Ponta Por\~{a},  Ponta Por\~{a} - MS, Brazil}

\noindent{$^3$S\~{a}o Carlos Institute of Physics, University of S\~{a}o Paulo, S\~{a}o Carlos - SP, PO Box 369, 13560-970, Brazil.\\Scientific Computing Group - http://scg.ifsc.usp.br}

\begin{abstract}
Dynamic texture is a field of research that has gained considerable interest from computer vision community due to the explosive growth of multimedia databases. In addition, dynamic texture is present in a wide range of videos, which makes it very important in expert systems based on videos such as medical systems, traffic monitoring systems, forest fire detection system, among others.
In this paper, a new method for dynamic texture characterization based on diffusion in directed networks is proposed. The dynamic texture is modeled as a directed network. The method consists in the analysis of the dynamic of this network after a series of graph cut transformations based on the edge weights. For each network transformation, the activity for each vertex is estimated. The activity is the relative frequency that one vertex is visited by random walks in balance. Then, texture descriptor is constructed by concatenating the activity histograms. 
The main contributions of this paper are the use of directed network modeling and diffusion in network to dynamic texture characterization. These tend to provide better performance in dynamic textures classification. Experiments with rotation and interference of the motion pattern were conducted in order to demonstrate the robustness of the method. The proposed approach is compared to other dynamic texture methods on two very well know dynamic texture database and on traffic condition classification, and outperform in most of the cases.
\end{abstract}

	\section{Introduction}
    
Dynamic texture is a spatio-temporal extension of image texture patterns. They are textures present in video or sequences of images that exhibit certain stationary properties in time \cite{din03}. Although simple and repetitive patterns could be considerate as dynamic texture, in general, they are characterized by non-rigid complex motions. These motions are guided by nonlinear and stochastic dynamics, for instance, the turbulent seawater or a fire blast \cite{dyntexfractal}.
Examples of dynamic textures are present in our daily lives (e.g., trees swaying smoke, waterfall, etc.) and in various areas of science, such as biology (e.g., evolution of bacterial colonies \cite{Zhang03082010,5291926}, tumor growth \cite{Celli3Dtumor2011,silva2012tumor,Wu2015}, tissue growth \cite{Khalil2009775}, etc) and material science (e.g., process of corrosion of metals and nanostructure \cite{Zimer20113193,Florindo20131694}).
	
Dynamic texture research appeared more than 30 years after static texture (textures in images), in the begging of the 90s \cite{Nelson:1992:QRM:167675.167694}. Moreover, only in the last decade that the attention to the area has increased. 
The dynamic texture is not just a trivial extension of the static textures, indeed, many issues arise when textures that change in time are analyzed. Besides the issues related to static textures, in dynamic textures, the patterns are now a combination in space and time. Some characteristics presented by dynamic textures include large amount of raw data, spatial and temporal regularity, and very little prior structure \cite{hu2011survey,Goncalves201651}.

In terms of application, dynamic texture methods are valuable tools in different areas such as industry, medicine, security, traffic engineering, among others. 
In these areas, there are various problems that can be modeled as dynamic texture. This makes the dynamic texture analysis an important research field.
The research in facial expression recognition using dynamic textures has proved to be more promise compared to static texture based approaches \cite{ZhaoLPB07}. 
In this way, fully automatic and real-time facial expression recognition could be used to help, for instance, biometric systems, psychological research, and human-computer interaction \cite{wang_micro_expression}. 
In medicine, dynamic texture methods were applied for automatic segmentation of liver in ultrasound, allowing the classification of hepatic structures including vasculature and liver parenchyma \cite{Milko2008}. 
Recently, works employed dynamic texture methods to identify the traffic conditions (i.e., light, medium, slow, and stopped traffic) and to support traffic monitoring system \cite{Chan08,Goncalves20134283}.
Besides that, there are many systems using dynamic texture methods. Some examples are forest fire detection systems \cite{Zhao2011,smoke_detection}, video retrieval systems \cite{Peteri2006,Ravichandran11}, recognition of human activity \cite{kellokumpu2008human}, among others.

The characterization of dynamic textures was much less studied than the static textures and has many challenges to overcome, such as: (i) extract features that integrates the combination of appearance and motion (most of the methods consider appearance and motion individually), (ii) invariance to transformations (e.g. rotation, scale and translation), (iii) the presence of complex texture patterns, (iv) multi-scale analysis (in the same way as the static texture, the dynamic textures can present different characteristics at different levels of scale, e.g. local and global features \cite{Krig2014}), (v) computational cost (since the analysis is over video, computational time is an important issue), among others. 
These challenges cause limitations in most of the literature approaches.

Several method categories have been used along the years for dynamic texture analysis. However, optical flow based methods are the most popular due to the low computational cost and the efficiency in motion analysis of videos. These methods reduce the dynamic texture video to the analysis of a sequence of motion patterns \cite{Goncalves20131163}.
In the first works, it was used the vector field of normal flow to characterize the global magnitude and direction of motion in dynamic textures \cite{Nelson:1992:QRM:167675.167694,Polana1997}. 
More recent methods in this category include the vector histogram of acceleration and speed \cite{lu2005dynamic}, features invariant to scale and rotation of normal flow \cite{Fazekas}, combination of normal flow characteristics and periodicity \cite{fluxoChin}, among others.
Although they are good for motion analysis, optical flow based methods have as main limitation, be not capable to analyze properly the appearance characteristics of videos.
On the other hand, spatio-temporal filtering based methods characterize the dynamic texture through decomposition at various scales using space-time filtering in order to explore local and global information of motion and appearance. 
In the initial study \cite{Wildes2000}, filters of oriented energy were used to analyze local space-time patterns. 
Later, other studies using wavelet transform \cite{Wu2001,Smith,Dubois2009} and spatio-temporal Gabor filter \cite{DBLP:journals/corr/abs-1201-3612} were used for dynamic texture characterization. 
Finally, a category of methods less popular is the geometric properties based methods. These methods use properties of moving contour surfaces (i.e., features based on the tangent plane distribution) to characterize the spatial and temporal domain \cite{OtsukaHorikoshi,Zhong2002}. 
    
In addition to the categories mentioned above, model-based methods have been proposed. These ones aim to build generative process models from the video and extract features from these models to characterize the dynamic texture \cite{chan2005classification,Chan08,ravichandran2013categorizing}. 
One of the most popular methods of this category is based on linear dynamical systems (LDS) model which inspired several researchers to use LDS to dynamic texture analysis. It explores the spatial and temporal regularities of dynamic textures for the characterization \cite{din03}.
Besides LDS models, fractal models have also been used to analyze dynamic textures such as dynamic fractal spectrum (DFS) \cite{dyntexfractal}, 3D oriented transform feature (3D-OTF) \cite{3d-otf} and wavelet domain multi-fractal spectrum (WMFS) \cite{wmfs}.
Recently in this category, have been proposed agent methods. They use walkers guided by deterministic rules to extract features of motion and appearance in dynamic textures \cite{Goncalves20131163,Goncalves20134283,Goncalves2011}.
Another recent method uses network models of the dynamic texture and extracts statistical measures from these networks \cite{Goncalves2015211}. 
This method achieved promising results in dynamic texture recognition, however, the dynamic texture was modeled into an undirected network and only three statistical measures of the degree distribution were used. We believe that other types of modeling and more robust measures can increase the performance of the dynamic texture recognition.

In this paper, we propose a new method for dynamic texture characterization based on diffusion in directed complex networks, following recent works that show that diffusion in directed networks may reveal clustering in the structure versus dynamics space \cite{diffusionCosta,Goncalves201651}. The main contributions of this paper are the modeling of dynamic texture as a directed network and the feature extraction using diffusion from networks.
	Complex networks is a term used for the study of graphs with mandatory presence of complex phenomena and their analysis by methods from mechanical statistics.
	Recently, complex networks have become an object of interest due to its flexibility and simplicity to represent complex systems. 
	Thus, several works have used networks to represent and characterize images of static textures with success \cite{Backes2013168, Backes200954, Goncalves201211818}. 
	In dynamic textures, the first approach that uses complex networks is reported in \cite{Goncalves2015211}. 
	
    In the proposed method, the dynamic texture is modeled as a regular directed network, which is characterized by the activity of the vertices computed by diffusion.
	To model the dynamic texture, each pixel is mapped into a vertex, which is connected to others vertices according to a given radius. 
	Then, the network dynamic is explored through thresholds that aim to highlight specific topological features from each dynamic texture class.
Given a transformed network, the activity of each vertex is estimated by random walks.
The activity is the relative frequency at which each vertex is visited in balance by random walks.
	In order to measure the topological characteristics, we propose the use of two histograms that associates the activity with the temporal and spatial degree of vertices.    
Experiments performed in two dynamic texture databases illustrate that our method achieved excellent performances in dynamic texture recognition. 
	In particular, in the UCLA-50, UCLA-9 and UCLA-8 databases, the proposed method achieved the highest correct classification rate compared to state-of-the-art methods. Besides that, experimental results on traffic condition classification have demonstrated the effectiveness of the proposed approach. Finally, experiments in synthetic textures showed that the proposed method is efficient in motion analysis and it is rotation invariant.

	This paper is organized as follows. 
	In Section 2 complex network theory is briefly discussed to provide motivation and background for the proposed extension. 
	A new method for dynamic texture characterization based on activity of directed complex networks is described in Section 3. 
	Section 4 presents the experimental setup that includes a description of the databases.  
	Experimental results on parameter evaluation, comparison with others methods and computational complexity analysis is presented in Section 5. 
	Finally, Section 6 concludes this paper.

	\section{Diffusion in networks}
	
	Graphs or networks? The two terms are related to the same object of studies that arose into the Mathematics as Graph Theory. Graph Theory started from the Euler's solution of the famous Konigsberg bridges problem in 1736 and, since then, has been studied in Mathematics and, after the 50s, in Computer Science. Into the last decades, the Graph Theory attracted the attention of physicists, which contributed to the field, incorporating analysis and methods that came from the Mechanical Statistics. The combination of graphs and mechanical statistics based analysis is known in the literature as Complex Networks or just Networks (where the presence of complex phenomena are not mandatory into the graphs) \cite{Newman:2010:NI:1809753,cohen2010complex}. In this work, let us use the term "network" because diffusion is associated to the complex networks field.
	
	A network can be defined as a pair $N=(V,E)$, where $V = \{ v_i\}$ is a set of vertices and $E$ is a set of edges $e_{v_i,v_j}$. For each edge, it is assigned a weight $e_{v_i,v_j} \in \Re$, which according to application represent lengths, costs, measures, etc.
The networks can be undirected (i.e., when the edges have no direction), or directed (i.e., an edge $e_{v_i,v_j}$ is directed when there is a direction from $v_i$ to $v_j$, thus, $v_i$ is called the tail and $v_j$ is called the head). In undirected networks, the degree $k_{v_i}$ of a vertex $v_i$ is the number of edges incident to the vertex. On the other hand, in directed networks each vertex $v_i$ has a in-degree $k_{v_i}^{in}$, corresponding to the number of ingoing edges (i.e. number of heads adjacent to vertex), and an out-degree $k_{v_i}^{out}$, equal to number of outgoing edges (i.e. number of tails adjacent to vertex).

	The diffusion in networks used in this work was proposed into the complex network field \cite{diffusionCosta}. In this method, the diffusion is estimated by means of traditional random walks. Thus, $N$ walks are initiated at each vertex and move according to the probability associated with the edge weights.
	For characterization of the topological structure of the network, it is estimated a histogram of activities. The activity of a vertex is the frequency in that it was visited by the walkers. Thus, the histogram is composed of the sum of the activity of all vertices with the same degree $k_{in}$.
	Recently, diffusion in networks has been used for static texture characterization \cite{Goncalves201651} through the histogram of the activities. Differently, in this work, we propose an approach to characterize dynamic textures. 

	\section{Dynamic texture using diffusion in networks}
    
\begin{figure}[!htbp]
		\centering
		\includegraphics[width=.9\columnwidth]{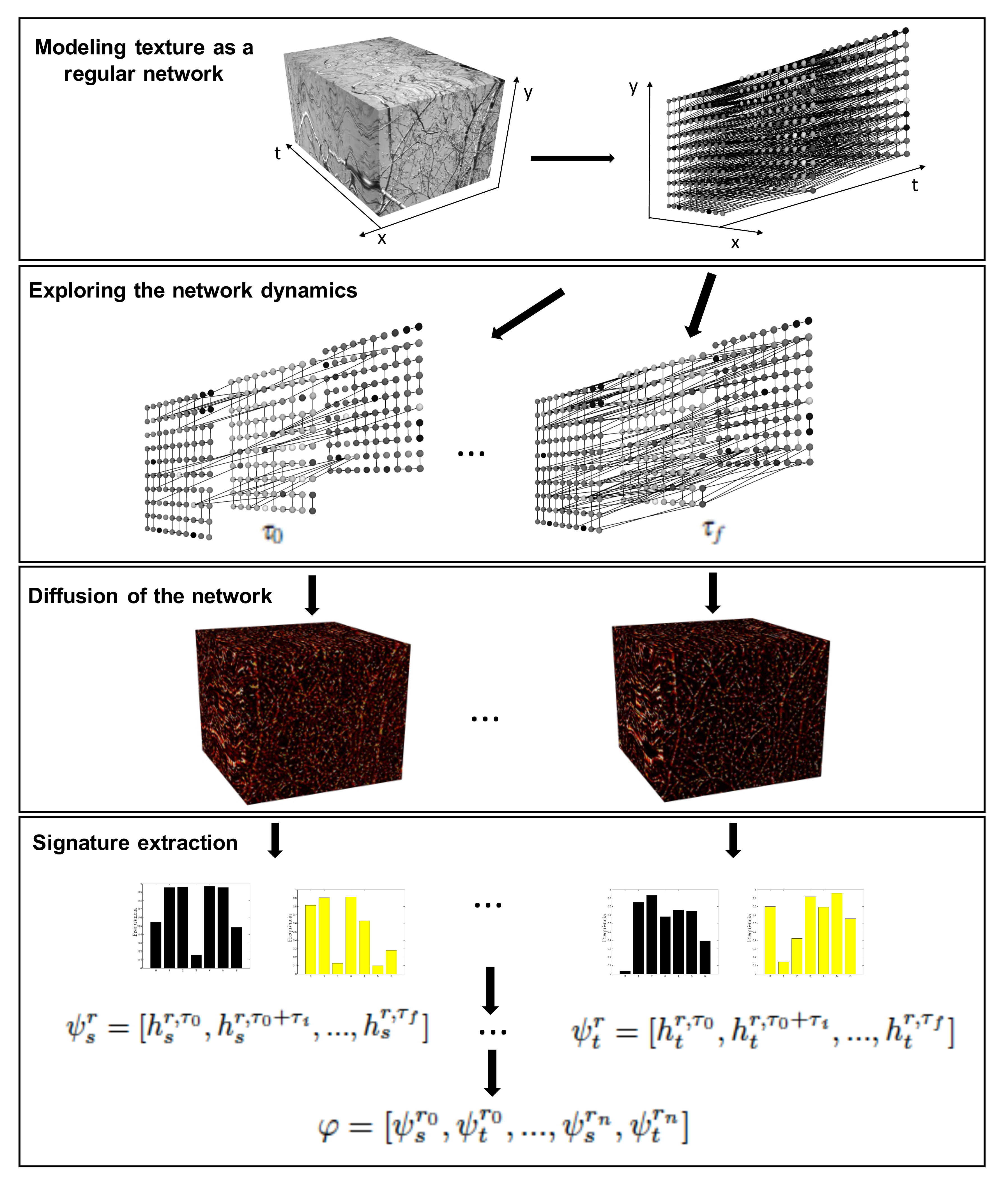}
		\caption{Illustration of the four steps of the proposed method for characterization of dynamic textures based on activity in directed networks.}
		\label{fig:activity_scheme}
	\end{figure}

	In this section, we describe the proposed method for dynamic texture characterization based on activity of directed networks.
    Basically, the proposed method can be described into four steps. 
    Figure \ref{fig:activity_scheme} summarizes the steps:
    
    \begin{itemize}
        \item \textbf{Modeling texture as a regular network:} first, the video of the dynamic texture is modeled as a regular directed network, where each pixel is considered a vertex. Each vertex is connected to other vertices if the Euclidean distance between the pixels represented by the vertices is lesser than or equal to a predefined value. A weight is defined as the gray intensity difference of the vertices.
        
        \item \textbf{Exploring the network dynamics:} the directed network is transformed by a cutting function imposed on the edges. This function removes edges whose weight is greater than a given threshold, highlighting specific characteristics of each texture pattern. 
        
        \item \textbf{Diffusion in network:} random walk starts in each vertex of the network obtained in step 2. The walker moves according to a random raffle, where the probability of visiting a vertex is associated with the edge weight. The activity of each vertex is the number of visits in balance.
        
        \item \textbf{Signature extraction:} characteristics of appearance and motion of the dynamic texture are considered to compose the feature vector. Thus, it is obtained an activity histogram associated to the temporal in-degree (i.e., edges that connect two vertices of different frames) and an activity histogram associated to the spatial in-degree (i.e., vertices in the same frame).

	\end{itemize}

	In the next subsections, each of the proposed method steps is described in details.
	
	\subsection{Modeling of dynamic textures in directed networks}
	
	In this approach, the video is considered as a three-dimensional matrix that contains all its frames. Thus, a pixel $i$ can be addressed as $i = (x_i, y_i, t_i)$, where $t$ is time or, in discrete terms, the frame that the pixel belongs, and $x, y$ are its spatial coordinates. In this model, the intensity of a pixel is given by $I(i) \in [0,255]$. 
	To model a video on a network $R = (V, E)$, each pixel $i$ is mapped to a vertex $v_i \in V$. Two vertices $v_i$ and $v_j$ associated to the pixels $i$ and $j$ are connected by an edge $e_ {v_i, v_j} \in E$, if the Euclidean distance $D(i,j)$ between $i$ and $j$ is lesser than or equal to a predefined value of radius $r$ \cite{Goncalves2015211}, according to,
	
	\begin{equation}
	\begin{split}
	e_{v_i,v_j} \in E \textnormal{ if } D(i,j) \leq r, \\
	D(i,j)=\sqrt{(x_i-y_j)^2 + (x_i-y_j)^2 + (t_i-t_j)^2}.
	\label{eq:dist}
	\end{split}
	\end{equation}

\begin{figure}[!htbp]
		\centering
		\includegraphics[width=.8\columnwidth]{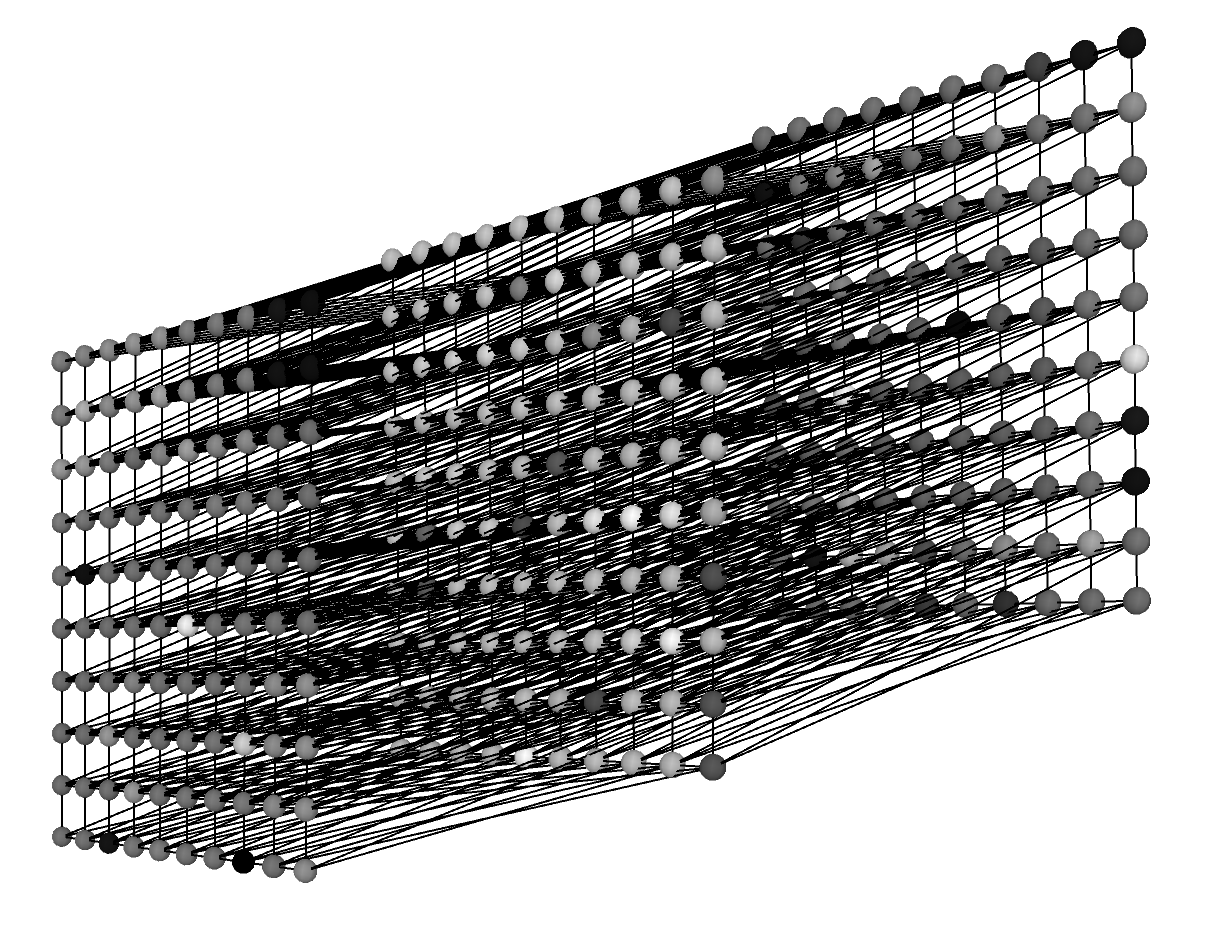}
		\caption{Example of regular network modeling a video with three frames.}
		\label{fig:regular_network}
	\end{figure}

	Figure \ref{fig:regular_network} shows a regular network obtained from a video with three frames in a given radius $r$. 
	Each vertex is connected with vertices of the same frame and different frames. 
	Therefore, each vertex has the same degree, except for those on the border.
	
	For each edge, it is assigned a weight  $w(e_ {v_i, v_j})$ according to the intensity difference between pixels corresponding to the vertices:
	
	\begin{equation} 
	\label{eq:weight}
	w(e_{v_i,v_j}) = 
	\left\{ \begin{array}{rcl}
	I(i) - I(j), & D(i,j) \leq r \\
	\textnormal{NaN}, & \textnormal{otherwise} 
	\end{array} \right.
	\end{equation}
	
 It is worth to mention that the weights of the edges are invariant to lighting changes if the change is the same for all frame video, i.e. if the lighting is constant  \cite{Goncalves2015211} (e.g. two videos of the same class captured with different lightings).
    Indeed, if the lighting conditions of the video are varying in each frame the method is not invariant to lighting changes. 
	
	Notice that according to Equation \ref{eq:weight}, negative weights are obtained when $I(i)<I(j)$. 
	In this study, it is considered only edges with positive weight. 
	Thus, an edge $e_{v_i, v_j}$ is directed from $v_i$ to $ v_j$, if $I(i)>I(j)$. 
	The modeling with directed networks is more suitable to capture topological features through the dynamic process, such as random walks, which differs from the undirected network used in previous dynamic texture works \cite{Goncalves2015211}.
	
\subsection{Threshold function}
    The network obtained in the previous step has regular behavior (if we do not consider the edge direction), i.e., all vertices have the same degree.  Therefore, information based on vertex degrees cannot characterize the network that models the dynamic texture.
	In this way, in this step, a transformation based on a series of edges cuts is applied over the network, revealing relevant properties of the dynamic texture. This approach is usually used in complex networks to extract information from the structure and dynamics \cite{costa2007characterization}. 
	Here, the transformation is performed applying a threshold function $\phi(\tau, E)$ over the set of edges of the network. 
    This function consists in selecting a set $E_\tau, E_\tau \subseteq E$, where each edge $e_{v_i,v_j} \in E_\tau$ has weight lesser than or equal to a given threshold $\tau$ (Equation \ref{eq:threshold}) \cite{Goncalves201211818, Backes2013168}.
	
	\begin{equation} \label{eq:threshold}
	E_\tau = \phi(\tau,E) = \{ e_{v_i,v_j} \in E \mid 0<w(e_{v_i,v_j})\leq \tau \}
	\end{equation}
	
\begin{figure}[!htbp]
		\centering
		\includegraphics[width=1.\columnwidth]{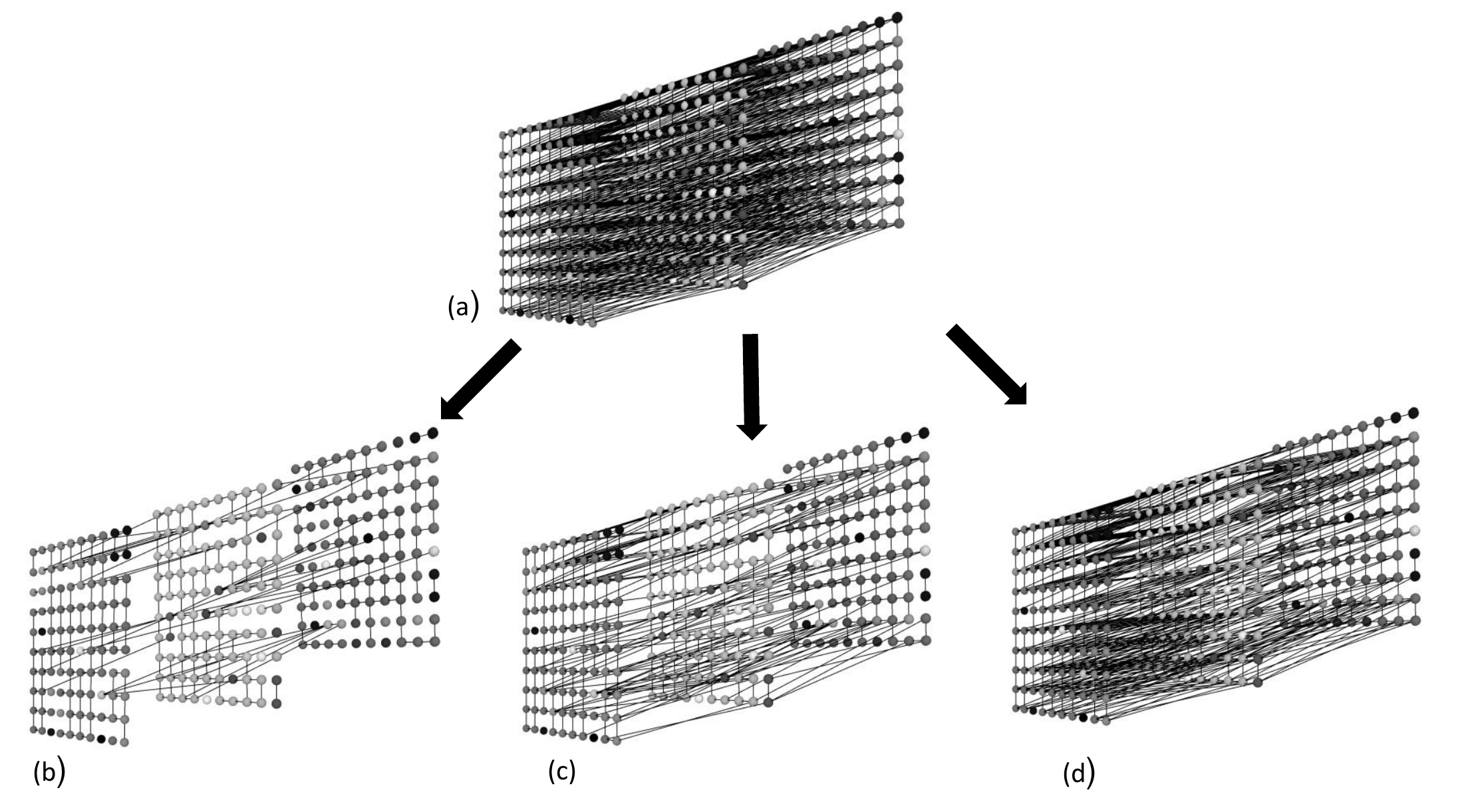}
		\caption{Examples of application of the function $\phi(\tau,E)$ on the edges of a regular network: (a) regular network, (b) $\tau=20$, (c) $\tau=50$, (d) $\tau=80$.}
		\label{fig:threshold}
	\end{figure}

	The new set of edges $E_\tau$ and the original set of vertices $V$ compose the new network $R_\tau={(V,E_\tau)}$, which represents an intermediate stage in the evolution of the network \cite{Backes2013168}.
    Figure \ref{fig:threshold} presents an example of the function  $\phi(\tau,E)$ applied on a regular network considering three values of $\tau$ (Fig. \ref{fig:threshold} (b) $\tau = 20$, Fig. \ref{fig:threshold} (c) $\tau = 50$ and Fig. \ref{fig:threshold} (d) $\tau =80$). The evolution of the network by applying the function $\phi(\tau,E)$ with different values of $\tau$ can be considered as a multi-scale analysis of the network \cite{GoncalvesSB10}. For low values of $\tau$ (Figure \ref{fig:threshold} (b)), it can analyze and describe micro texture details, as $\tau$ increases, global details are highlighted. Thus, the proposed method applies a set of thresholds $\tau$, $\tau \in \Gamma $ on the original network in order to extract features from its dynamics in a multi-scale approach. 
	
\subsection{Activity estimation in dynamic textures}
    
	To characterize a $\tau$-scaled network transformed by the threshold function, the activity of each vertex is estimated by random walks.
    Consider a walker that is in the vertex $v_i$, the next step is to visit a vertex $v_j$ with probability:
	
	\begin{equation}\label{eq:walk}
	p(v_i,v_j) = \frac{w(e_{v_i,v_j})}{\sum_{v_k \in V}^{} w(e_{v_i,v_k})}.
	\end{equation}

	The walker conducts the walk according to Equation \ref{eq:walk} until it visits a vertex without outgoing edge or the length of the walking is greater than a given threshold  $L$. 
	The activity $\alpha(v_i)$ of the vertex $v_i$ is defined as the number of visits received during the walks.
	To estimate more accurately the activity, $M$ walks are started at each vertex of the network \cite{Goncalves201651}. 
    According to experimental evaluations in this paper, we define $M=50$.
	
\begin{figure}[!htbp]
		
		\centering
		\subfigure[Dynamic texture ]{\includegraphics[width=0.45\textwidth]{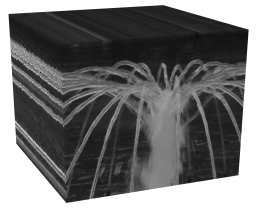}}\\
		\subfigure[$r=\sqrt{2}, \tau=30$.]{\includegraphics[width=0.45\textwidth]{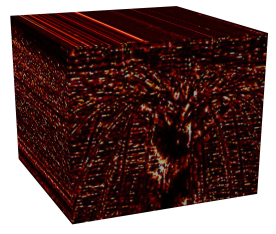}}
		\subfigure[$r=\sqrt{2}, \tau=120$.]{\includegraphics[width=0.45\textwidth]{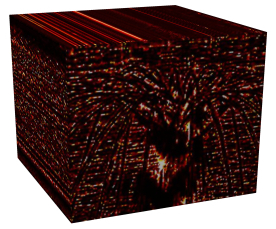}}
		\subfigure{\includegraphics[width=0.07\textwidth]{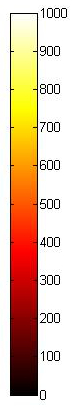}}\\
		\subfigure[$r=\sqrt{5}, \tau=30$.]{\includegraphics[width=0.45\textwidth]{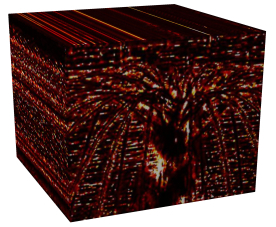}}    
		\subfigure[$r=\sqrt{5}, \tau=120$.]{\includegraphics[width=0.45\textwidth]{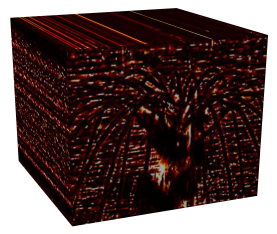}}
		\subfigure{\includegraphics[width=0.07\textwidth]{paleta.jpg}}
		\caption{Activity of the pixels/vertices of the dynamic texture, with various values of radius $r$ and thresholds $\tau$.}
		\label{fig:activity}
	\end{figure}

	Figure \ref{fig:activity} shows images of the activity of pixels/vertices of videos modeled with different radius $r$ and thresholds $\tau$.
	Note that, the activity of the pixels reflects the main properties of the dynamic texture, maintaining the appearance and motion characteristics. 
	Pixels in heterogeneous regions tend to have high activity due to the weight of their edges be higher than those in homogeneous regions. For example, if the edge weight $w(e_{v_i,v_j})$ between vertex $v_i$ and $v_j$ is 255 (the highest possible weight), the probability of the agent be attracted to visit the pixel $p_j$ is high.

\begin{figure}[!htbp]
		
		\centering
		\subfigure{\includegraphics[width=0.45\textwidth]{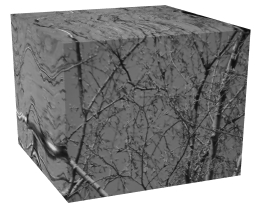}}
		\subfigure{\includegraphics[width=0.45\textwidth]{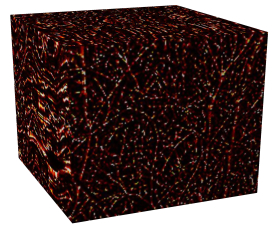}}
		\subfigure{\includegraphics[width=0.07\textwidth]{paleta.jpg}}\\
		\subfigure{\includegraphics[width=0.45\textwidth]{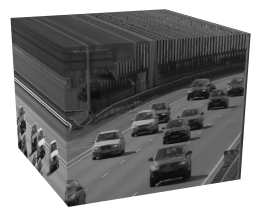}}    
		\subfigure{\includegraphics[width=0.45\textwidth]{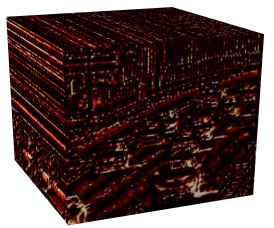}}
		\subfigure{\includegraphics[width=0.07\textwidth]{paleta.jpg}}\\
		\subfigure{\includegraphics[width=0.45\textwidth]{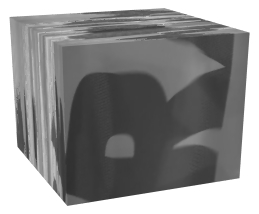}}
		\subfigure{\includegraphics[width=0.45\textwidth]{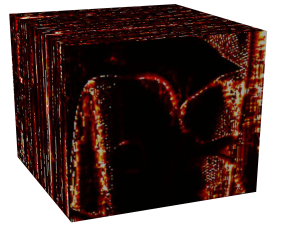}}
		\subfigure{\includegraphics[width=0.07\textwidth]{paleta.jpg}}\\
		\caption{Activity of the pixels/vertices of three dynamic textures. The network was build with radius $r=\sqrt{2}$ and thresholds $\tau=120$.}    
		\label{fig:video_activity}
	\end{figure}

	In Figure \ref{fig:video_activity}, the activity of pixels/vertices to different videos modeled with $r=\sqrt{2}$ and $\tau=120$ is shown.
	It can be seen that the activity can characterize different types of dynamic textures, capturing characteristics of motion and appearance. 
	
	\subsection{Composing the feature vector by the activity histogram}
	
	The activity of the vertices describes important properties of the dynamic texture. In general, to characterize a static texture, a histogram of activity that correlates the activity of the vertex and its in-degree is used \cite{Goncalves201651}. 
	However, as mentioned before, in dynamic textures one of the challenges is to describe the appearance and motion properties.
	This paper proposes to correlate the activity $\alpha(v_i)$ of the vertex $v_i$ with their spatial in-degree $k_{s}^{in}(v_i)$ and temporal in-degree $k_{t}^{in}(v_i)$. 
	The spatial in-degree $k_{t}^{in}(v_i)$ is the number of incoming edges from vertices that are in the same frame:
	
	\begin{equation}\label{eq:grau_m}
	k_{s}^{in}(v_i) = \sum_{v_j \in V}^{} \left\{\begin{array}{rcl}
	1,& e_{v_i,v_j} \in E \textnormal{ and } t_i=t_j  \\ 
	0,& \textnormal{otherwise}
	\end{array}\right.
	\end{equation}
	
	On the other hand, the temporal in-degree $k_{t}^{in}(v_i)$ of a vertex $v_i$ is the number of incoming edges from vertices in different frames:
	
	\begin{equation}\label{eq:grau_t}
	k_{t}^{in}(v_i) = \sum_{v_j \in V}^{} \left\{\begin{array}{rcl}
	1,& e_{v_i,v_j} \in E \textnormal{ and } t_i \neq t_j  \\ 
	0,& \textnormal{otherwise}
	\end{array}\right.
	\end{equation}
	
	The joint distribution $S^{r,\tau}(s,t)$ is obtained by combining the spatial and temporal in-degree. This distribution is a sum of activities of the vertices with spatial in-degree $s$ and temporal in-degree $t$ (Equation \ref{eq:dist_joint}). The distribution is normalized by the number of vertices of the network $N=W\times H \times T$.
	
	\begin{equation}
	S^{r,\tau}(s,t) = \frac{1}{N} \sum_{v_i \in V}^{} \left\{\begin{array}{rcl}
	\alpha^{r,\tau}(v_i), & k_{s}^{in}(v_i) = s \textnormal{ and } k_{t}^{in}(v_i) = t   \\ 
	0,& \textnormal{otherwise}
	\end{array}\right.
	\label{eq:dist_joint}
	\end{equation}
where $r$ is the radius and $\tau$ is the threshold used to build the network.
	
\begin{figure}[!htbp]
		\centering
		\includegraphics[width=1.\columnwidth]{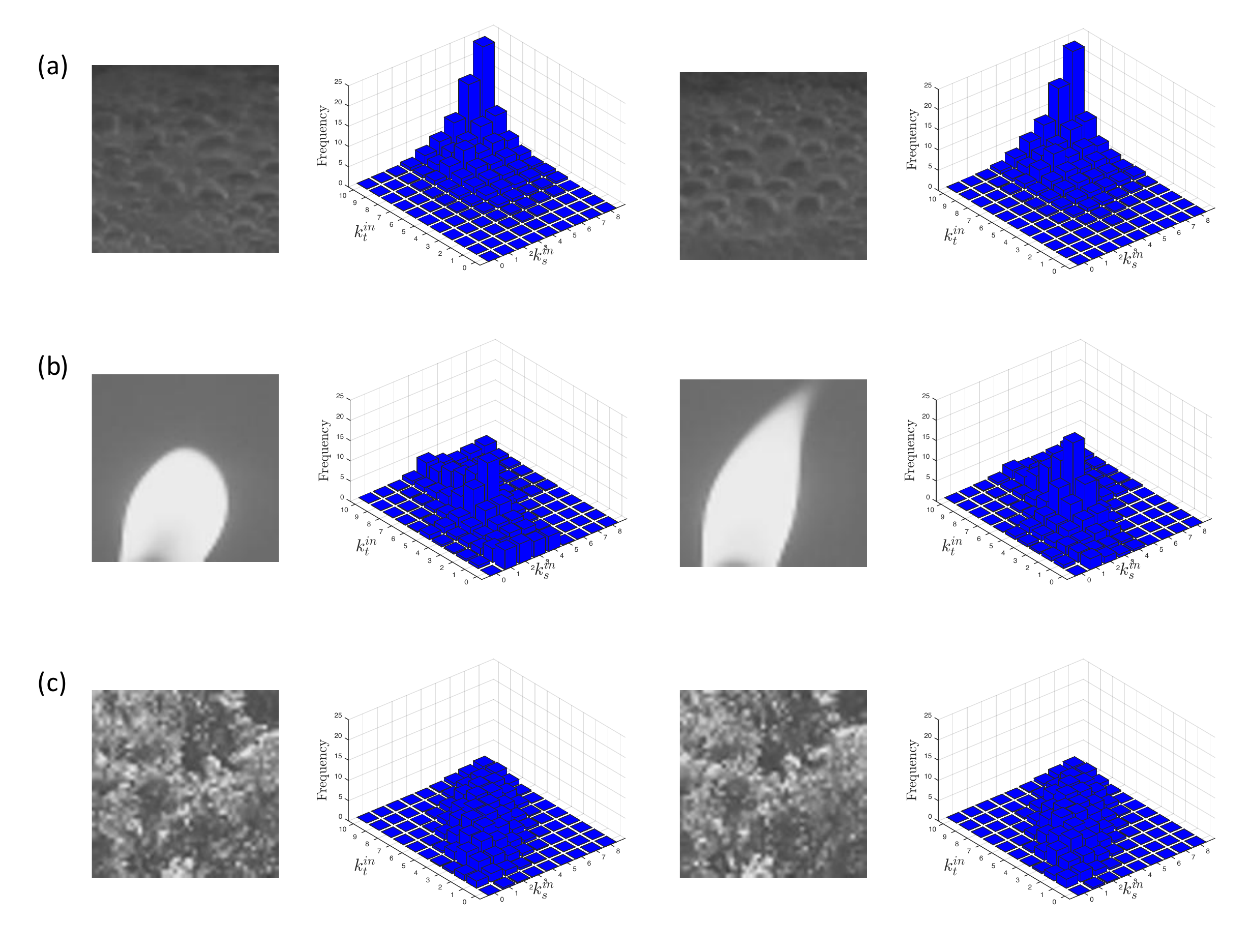}
		\caption{Examples of joint distribution obtained of three dynamic textures (illustrated by the first frame). The dynamic textures were modeled in networks with $r=\sqrt{2}$ and $\tau=30$.}
		\label{fig:dist_conj}
	\end{figure}

	This joint distribution has properties that allow describing the dynamic textures present in the video, which can be noticed in Figure \ref{fig:dist_conj} by the six examples. The classes of dynamic textures shown in figure are: boiling water (Figure \ref{fig:dist_conj}(a)), candle (Figure \ref{fig:dist_conj}(b)) and foliage (Figure \ref{fig:dist_conj}(c)).
	The dynamic textures were modeled using a directed network with $r=\sqrt{2}$ and $\tau=30$, and, as can be observed, each dynamic texture class presents a different joint distribution. The boiling water class is most homogeneous compared to others classes, thus, the weight of the edges are not high. That causes the activity be larger in vertices with high in-degree, as can be seen in the joint distribution of Figure \ref{fig:dist_conj}(a). In contrast, in Figure \ref{fig:dist_conj} (b) and (c) the vertices with a high in-degree do not have high activity, which reflects the characteristics of the dynamic texture.
	
	However, due to the large amount of information present in the joint distribution, two histograms are calculated, in order to reduce the feature vector size. 
	On one hand, the histogram $h^{r,\tau}_s(j)$  correlates the activity $\alpha(v_i)$  and the spatial in-degree $k^{in}_s$  (Equation \ref{eq:hist_m}) and on the other hand, the histogram $h^{r,\tau}_t(j)$ correlates the activity $\alpha(v_i)$  and the temporal  in-degree  $k^{in}_s$ (Equation \ref{eq:hist_t}).
	
	\begin{equation}\label{eq:hist_m}
	h^{r,\tau}_s(j) = \sum_{v_i \in V}^{} \left\{\begin{array}{rcl}
	\alpha^{r,\tau}(v_i),& k_{s}^{in}(v_i)==j  \\ 
	0,& \textnormal{otherwise}
	\end{array}\right.
	\end{equation}
	
	\begin{equation}\label{eq:hist_t}
	h^{r,\tau}_t(j) = \sum_{v_i \in V}^{} \left\{\begin{array}{rcl}
	\alpha^{r,\tau}(v_i),& k_{t}^{in}(v_i)==j  \\ 
	0,& \textnormal{otherwise}
	\end{array}\right.
	\end{equation}
	
    To discuss the histograms of Equations \ref{eq:hist_m} and \ref{eq:hist_t}, consider the following interpretation. 
    It is expected that vertices with high in-degrees have greater probability of it being visit during the walks.
However, this does not always occur due to the weight of the edges that are considered for during the walking process. 
These weights directly reflecting the properties of dynamic textures on the network, since they are the difference of intensity between the pixels. 
    Besides that, a vertex with low temporal in-degree can also have high activity if its spatial in-degree is high. This situation reflects more the appearance of the texture than the motion. The opposite situation can also occur. 
    Therefore, the histograms contain information correlated with the motion and appearance of the dynamic texture.    
	
\begin{figure}[!htbp]
		\centering
		\subfigure[First frame of the dynamic texture]{\includegraphics[width=0.84\textwidth]{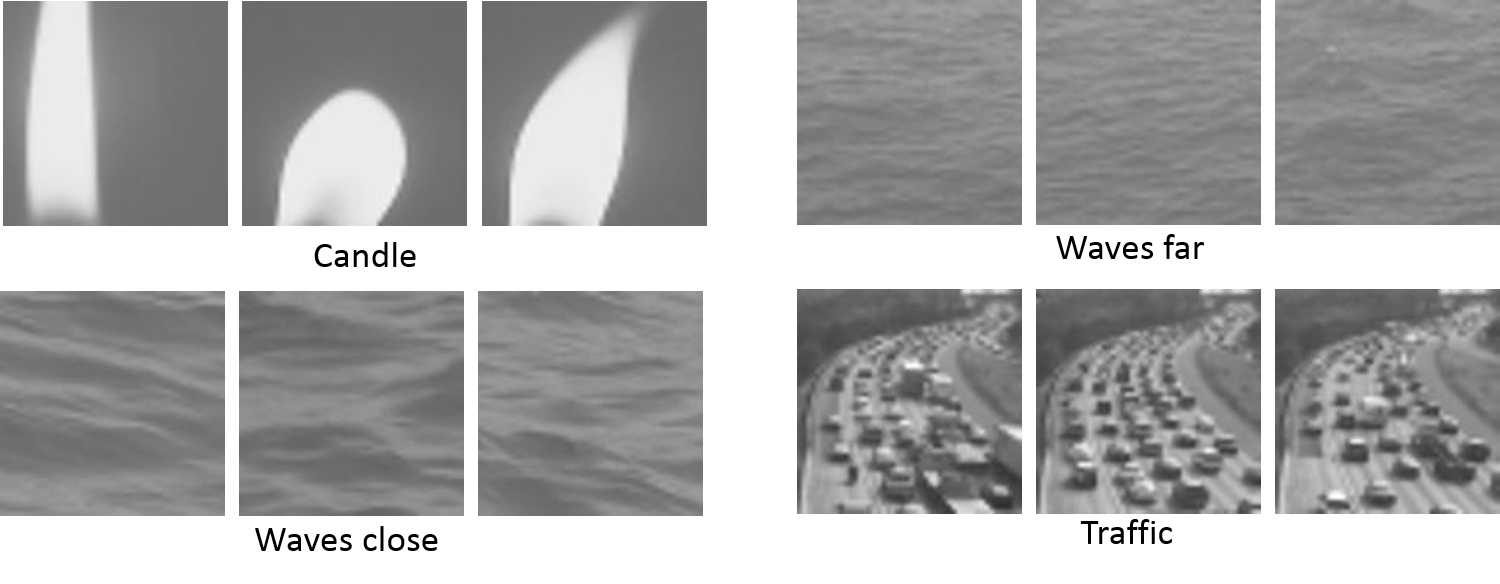}}\\
		\subfigure[Activity histogram $h^{r,\tau}_s$]{\includegraphics[width=0.45\textwidth]{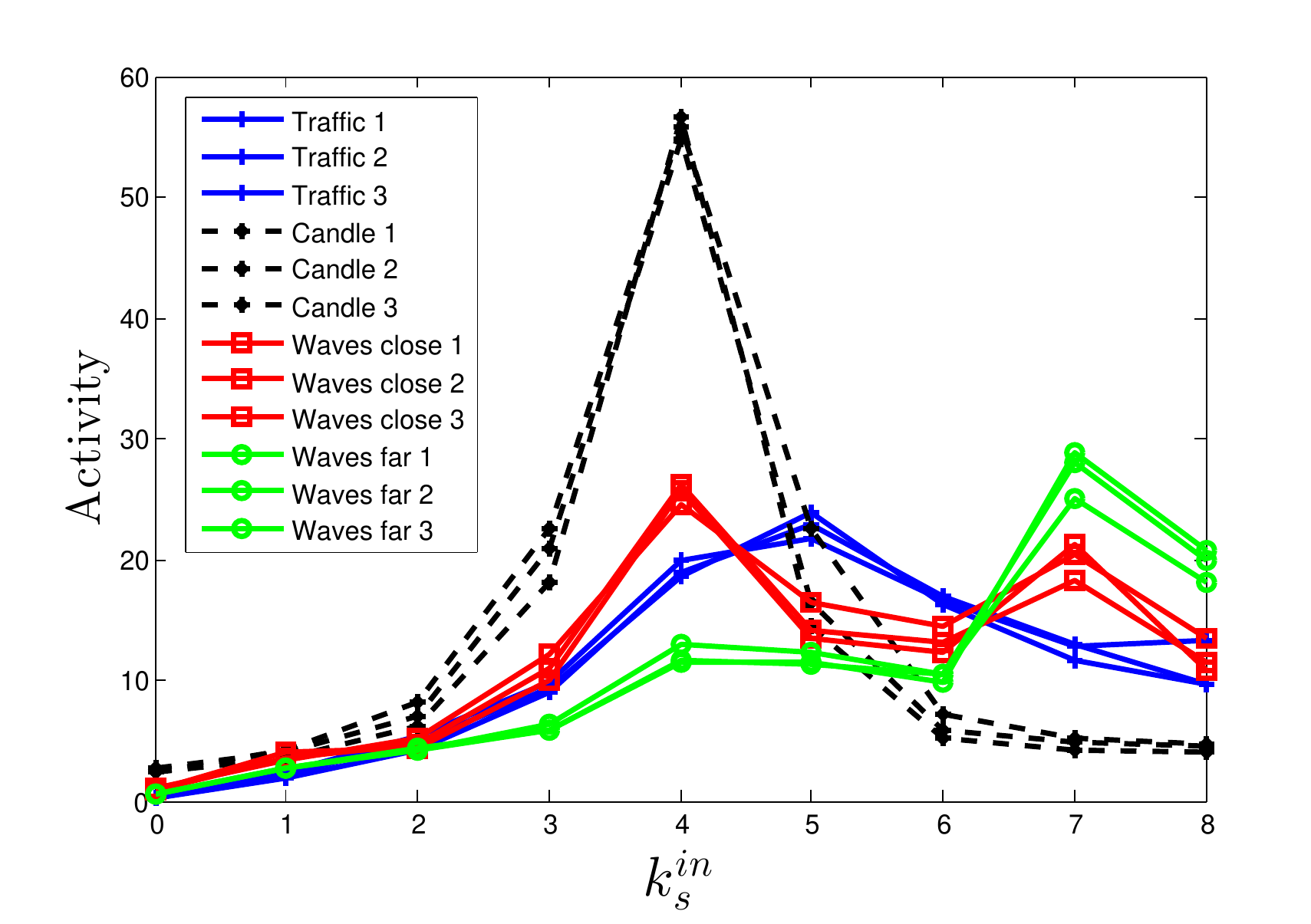}}
		\subfigure[Activity histogram $h^{r,\tau}_t$]{\includegraphics[width=0.45\textwidth]{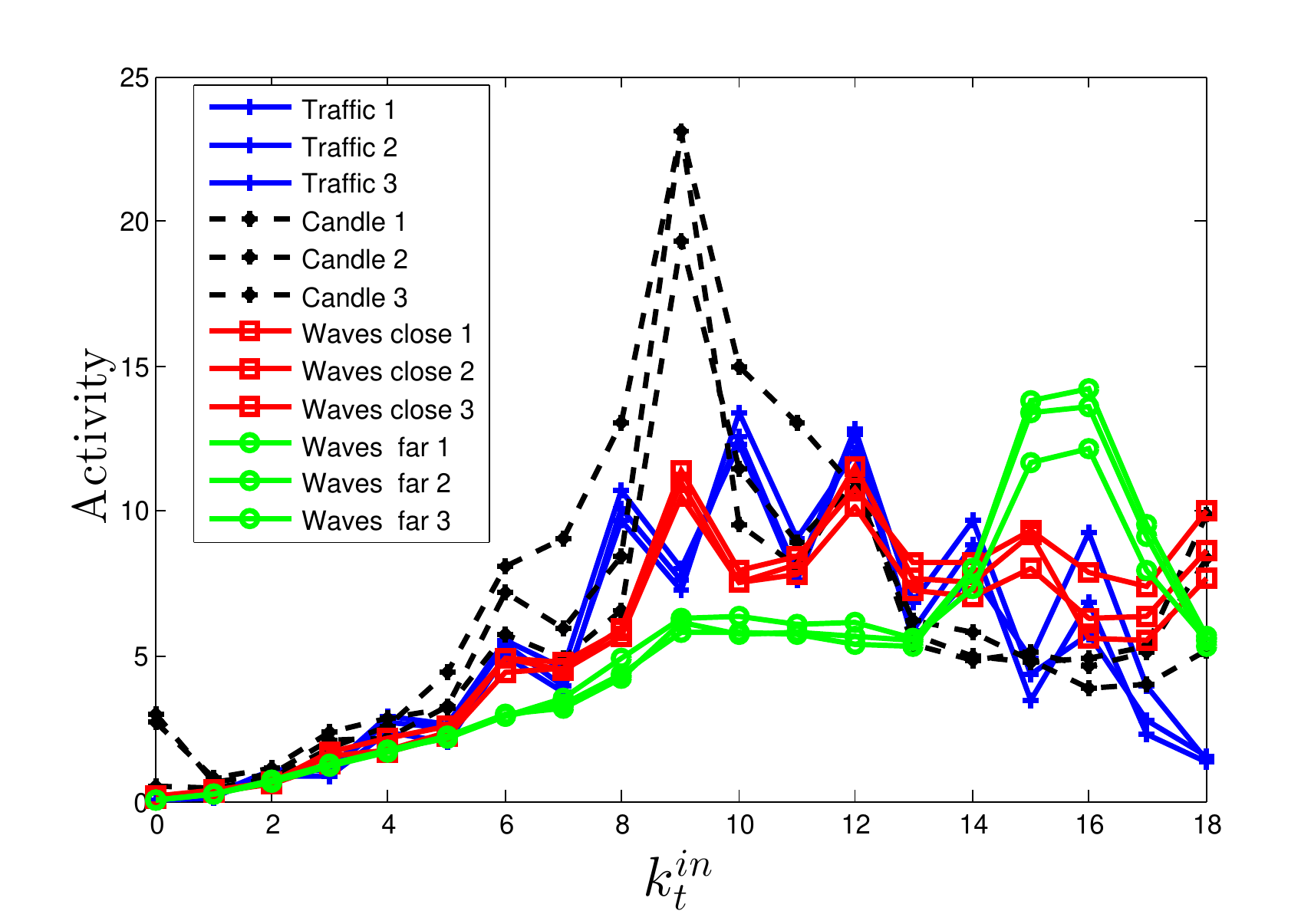}}
		\caption{Activity histogram associated to spatial in-degree $h^{r,\tau}_s(j)$ and temporal  $h^{r,\tau}_t(j)$ for 4 classes of dynamic texture. The dynamic textures were modeled in networks with $r=\sqrt{3}$ e $\tau=100$.}    
		\label{fig:hist_ativ}
	\end{figure}

	Figure \ref{fig:hist_ativ} shows examples of activity histograms $h^{r,\tau}_s(j)$ and $h^{r,\tau}_t(j)$  with $r=\sqrt{3}$ and $\tau=100$ for four different classes of dynamic texture. 
	The classes are burning candles, ocean waves away, ocean waves near and heavy traffic of a highway. 
	Considering the plot of the histogram $h^{r,\tau}_s(j)$ (Figure \ref{fig:hist_ativ}(b)) and  $h^{r,\tau}_t(j)$ (Figure \ref{fig:hist_ativ}(c)), it can be notice that each class of dynamic texture presents a different histogram, which demonstrate the potential to recognize the dynamic texture, even for classes that are similar as the case of the waves (far and close).

	For a dynamic texture multi-scale analysis we consider the histograms for different threshold values $\tau$, $\tau \in \Gamma$.
	The set $\Gamma$ is composed by threshold values that are in the range $\tau_0 \leq \tau \leq \tau_f$ and incremented by the constant $\tau_i$.
	Therefore, the feature vector of the histogram $\psi^r_s$ considering networks transformed with the same radius is given by:
	
	\begin{equation}
	\psi^r_s = [h^{r,\tau_0}_s, h^{r,\tau_0+\tau_i}_s,...,h^{r,\tau_f}_s]
	\end{equation}
	
	Similarly, the histogram $\psi^r_t$ is given by:
	
	\begin{equation}
	\psi^r_t = [h^{r,\tau_0}_t, h^{r,\tau_0+\tau_i}_t,...,h^{r,\tau_f}_t]
	\end{equation}
	
	Finally, to describe motion and appearance characteristics, the final vector $\varphi$ consists of the concatenation of vectors $\psi^r_s$ and $\psi^r_t$ for different values of radius $r$:
	
	\begin{equation}
	\varphi = [\psi^{r_0}_s,\psi^{r_0}_t,...,\psi^{r_n}_s,\psi^{r_n}_t].
	\end{equation}

	\section{Experimental setup} \label{sec:experiment}
    In this section, we describe the experimental setup used in the experiments performed with the proposed approach for recognizing dynamic textures based on diffusion in directed networks.
    In the experiments, a feature vector is extracted from each video of the database and it is classified using k-nearest neighbor, with $k = 1$. The choice of this classifier is due to its simplicity, highlighting the importance of the features extracted by the methods without any parameter optimization \cite{amancio2014systematic}. 
    For the evaluation of the proposed method, we have adopted an experimental setup similar to \cite{ravichandra_UCLA89,Ghanem:2010:MMD:1888028.1888046}. 
 To separate the training and test sets in Dyntex++ and UCLA-50 databases is used the $k-$fold cross-validation scheme with $10-$fold and $4-$fold in each database, respectively.
 The correct classification rate (CCR) is reported as the average performance of 10 experimental trials.
In UCLA-8 and UCLA-9 databases for the training set, half of the sequences are randomly selected from each class, and for the test set, the remaining half is used. 
In these two databases, the experiments are repeated 20 times and the average performance (CCR) and standard deviation of all trials are reported.
For the Traffic database, we have used a $10-$fold cross-validation scheme with 10 experimental trials.
    Next, we describe the three databases and its versions used in the experiments.

\begin{figure}[!htbp]
		\centering
		\includegraphics[width=.5\columnwidth]{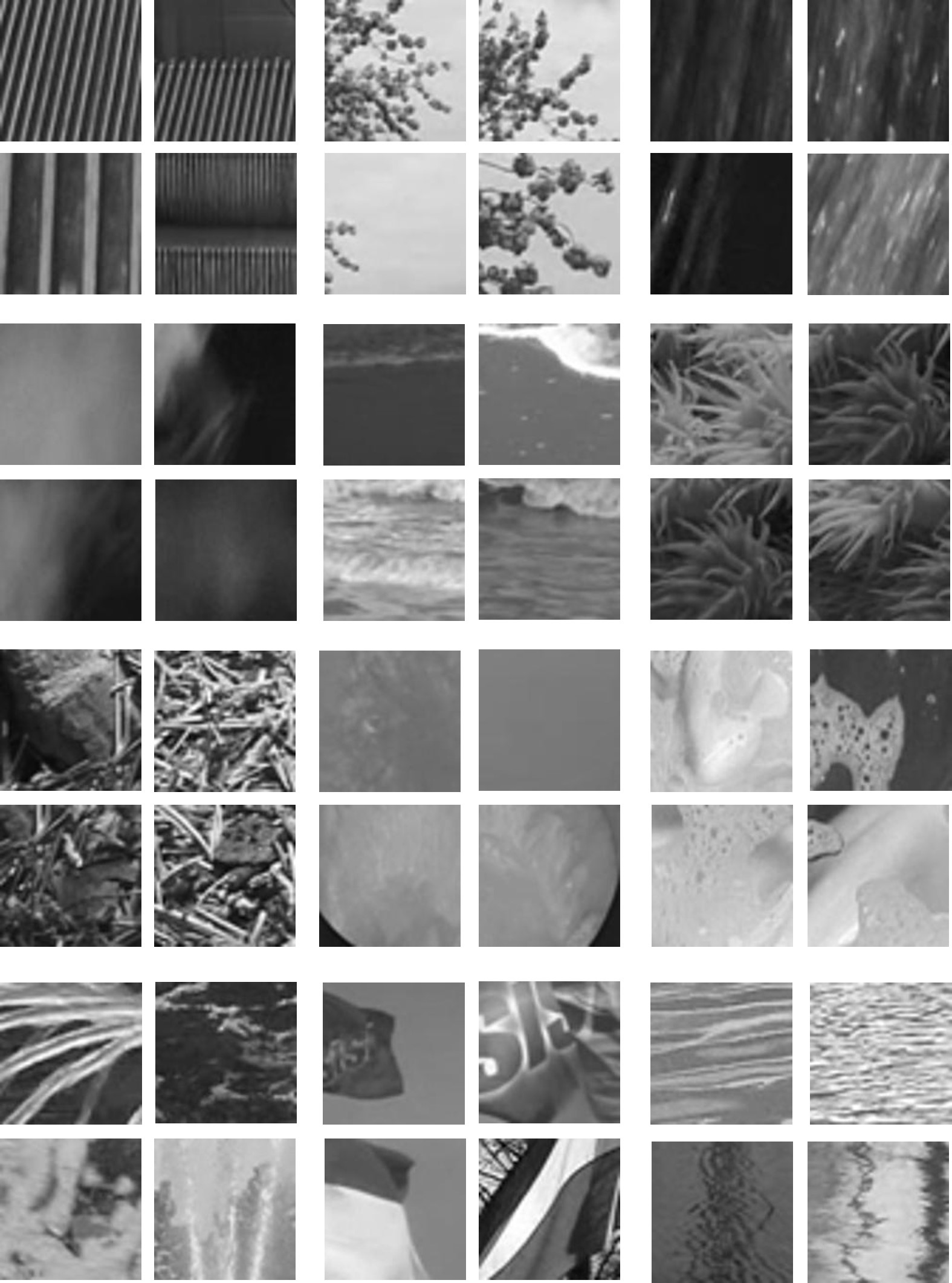}
		\caption{Examples of dynamic textures, illustrated by the first frame of the Dyntex++ database.}
		\label{fig:base_dyntex++}
	\end{figure}

\begin{figure}[!htbp]
		\centering
		\includegraphics[width=.5\columnwidth]{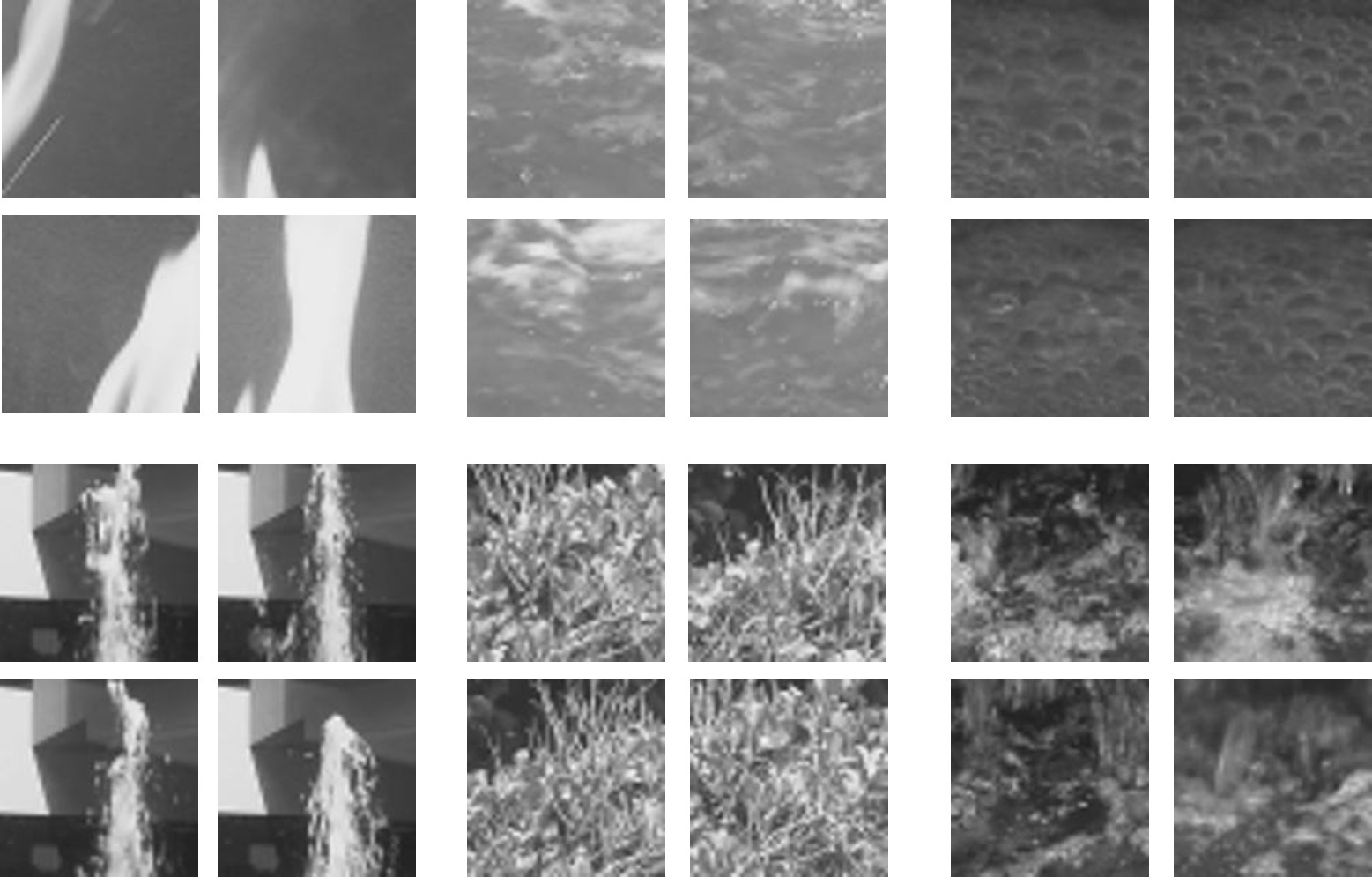}
		\caption{First frame of dynamic textures of UCLA database.}
		\label{fig:base_ucla}
	\end{figure}

\begin{figure}[!htbp]
		\centering
		\includegraphics[width=.5\columnwidth]{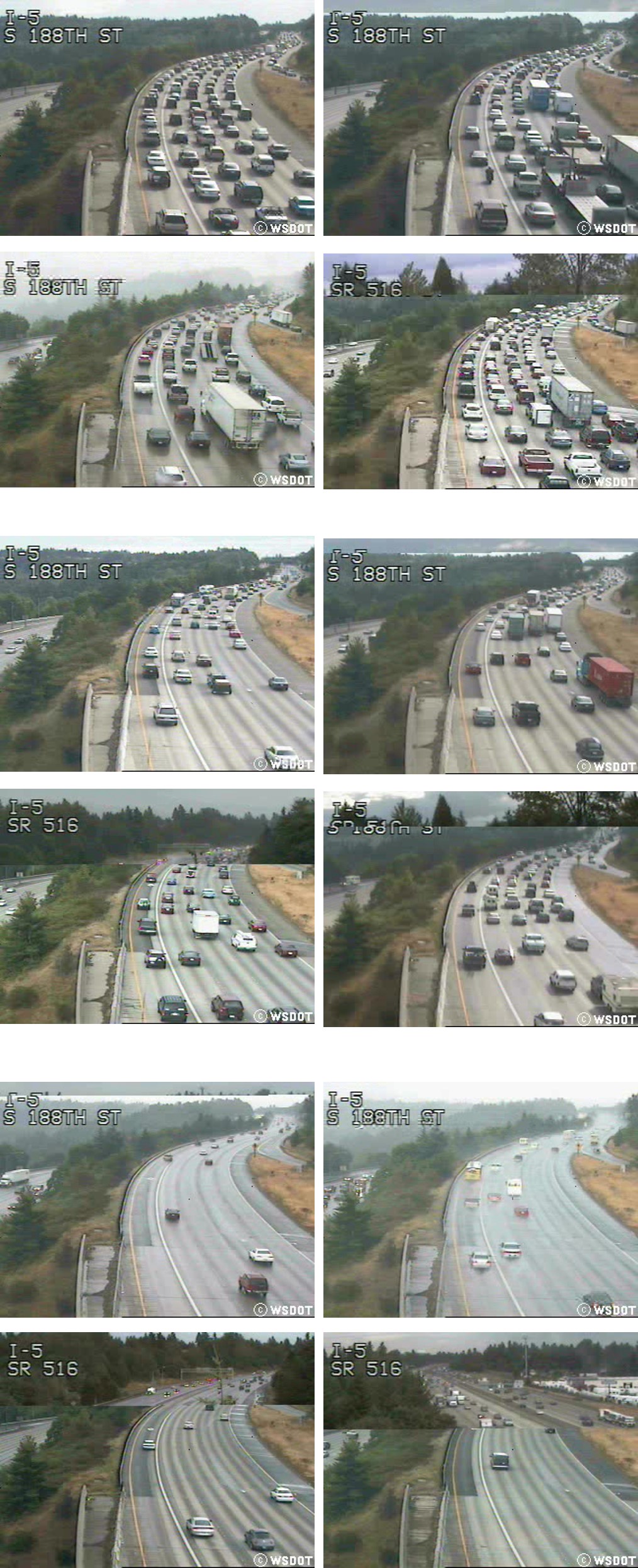}    
		\caption{Traffic database: examples of the first frame of three classes (heavy, medium and light traffic)}    
		\label{fig:base_traffic}
	\end{figure}

	\begin{itemize}
		\item Dyntex++ \cite{Ghanem:2010:MMD:1888028.1888046}: this database consists in 3600 videos of dynamic textures, divided into 36 classes with 100 samples each. Examples of dynamic textures presents into Dyntex++ are boiling water, river water, ant colony, smoke, among others. Figure \ref{fig:base_dyntex++} shows the first frame of some classes of Dyntex++.
		
		\item UCLA \cite{chan2005probabilistic,990925}: the UCLA database is a very popular benchmark for dynamic textures. The database contains 200 samples divided into 50 dynamic textures classes (UCLA-50), with 4 samples per class. In Figure \ref{fig:base_ucla}, we show the first frame of some dynamic textures presented in the database.
        We also have used two different variations proposed in \cite{ravichandra_UCLA89}. The first version reorganize the UCLA-50 database in 9 classes (UCLA-9) in order to combine videos that are taken from different viewpoints.
        The 9 classes are: boiling water (8), fire (8), flower (12), fountains (20), plants (108), sea (12), smoke (4), water (12) and waterfall (16). The number of samples per class is reported in parentheses. The second version (UCLA-8) discards the plant class to balance the number of samples per class.

		\item Traffic \cite{chan2005classification}: this database is composed by traffic videos, divided into three categories: light, medium and heavy traffic. The database consists of 254 videos in total with different traffic patterns and weather conditions. Some examples of videos of the three classes are shown in Figure \ref{fig:base_traffic}.
		
	\end{itemize}
	
	\section{Results and discussion}
    In this section, we evaluate the performance of the proposed method in four analyzes. 
    First, we present the influence of the method parameters in the three databases described in Section \ref{sec:experiment}. 
    Then, we evaluate the method behavior in characterizing motion features.
    We also present the performance of the proposed method in dynamic texture classification and compared with the other literature methods.
    Finally, it is performed an analysis of the computational complexity and time processing.
	
	\subsection{Parameter evaluation}
	
	In order to evaluate the proposed method, a parameter analysis is conducted to understand its influence on the dynamic texture classification task. The parameters are the initial threshold $\tau_0$, the incremental threshold $\tau_i$, the final threshold $\tau_f$ and the radius $r$ to model the dynamic texture as a directed network. Next, the classification results for different parameter settings are presented.

\begin{figure}[!htbp]
		\centering
		\subfigure[Initial threshold ]{\includegraphics[width=0.47\textwidth]{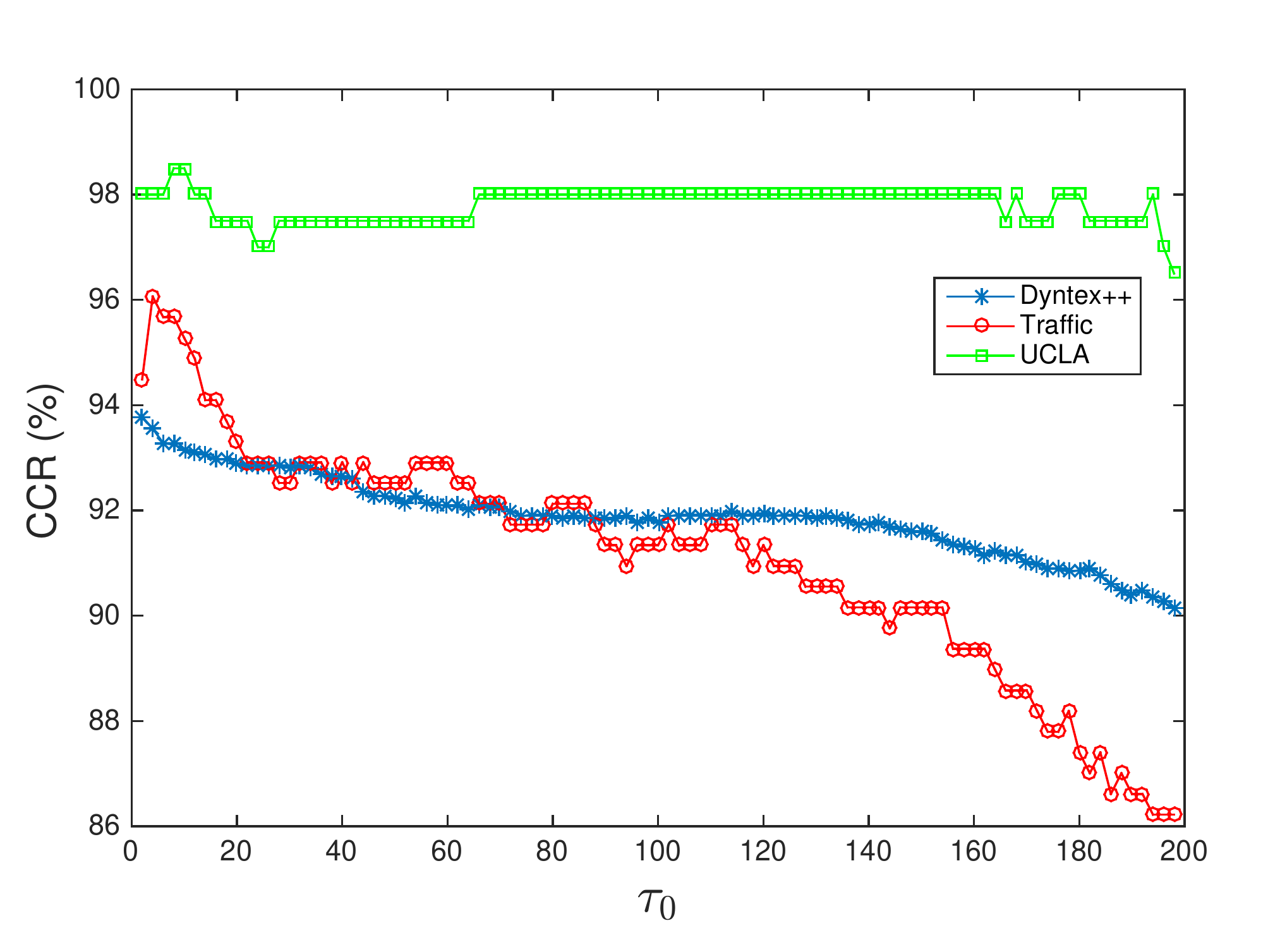}}
		\subfigure[Incremental threshold ]{\includegraphics[width=0.47\textwidth]{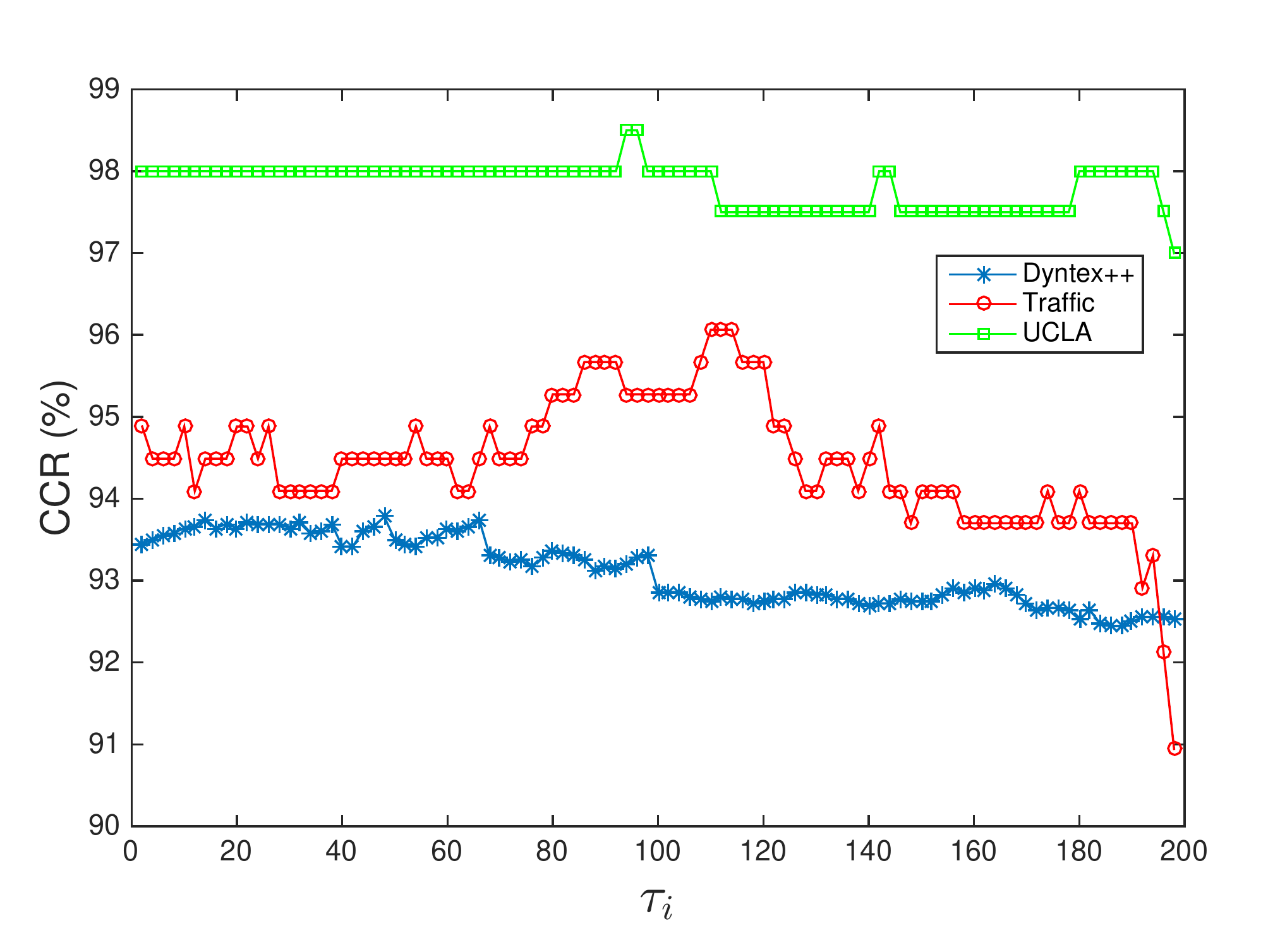}}\\
		\subfigure[Final threshold]{\includegraphics[width=0.45\textwidth]{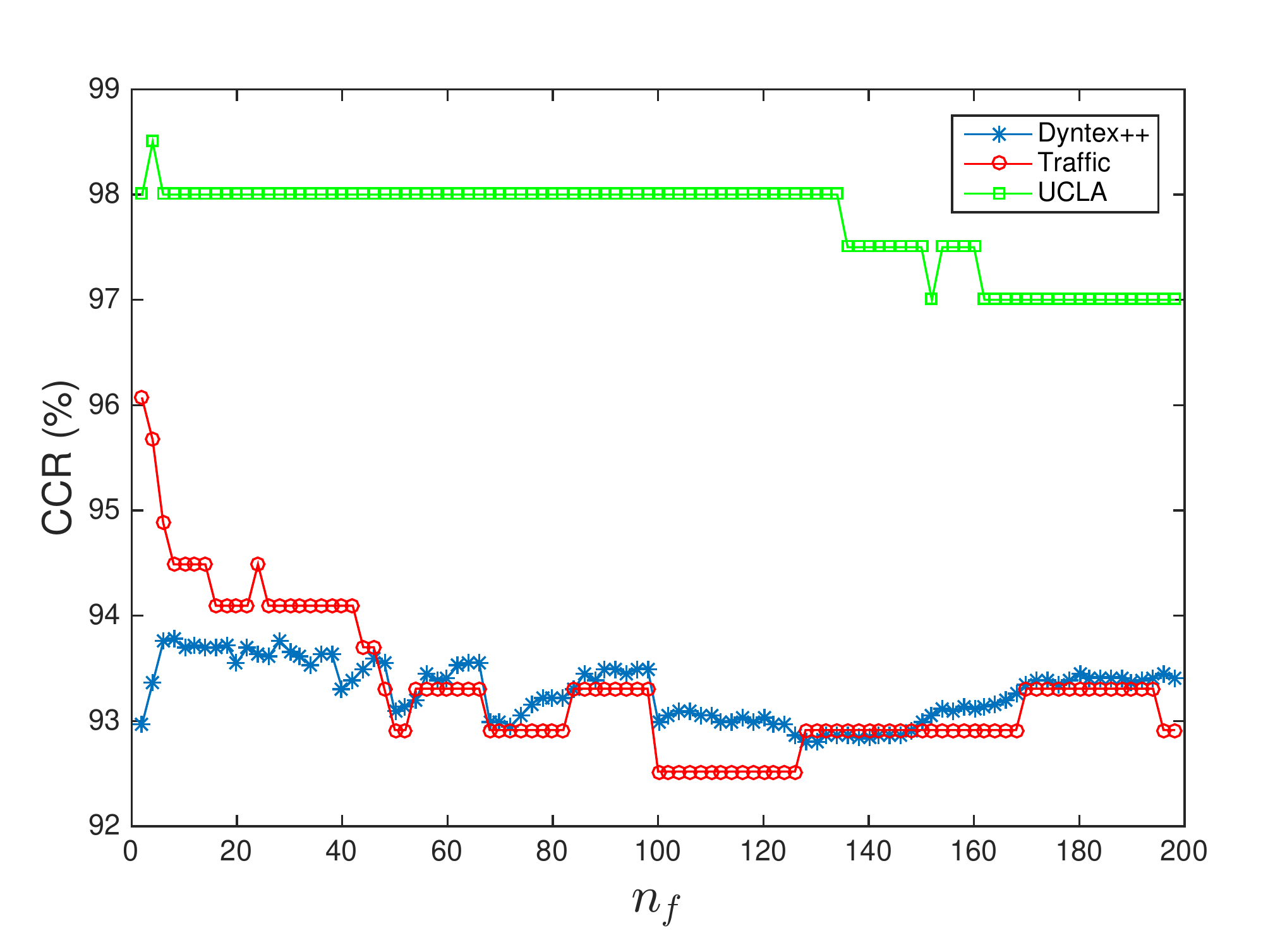}}
		\subfigure[Radius]{\includegraphics[width=0.45\textwidth]{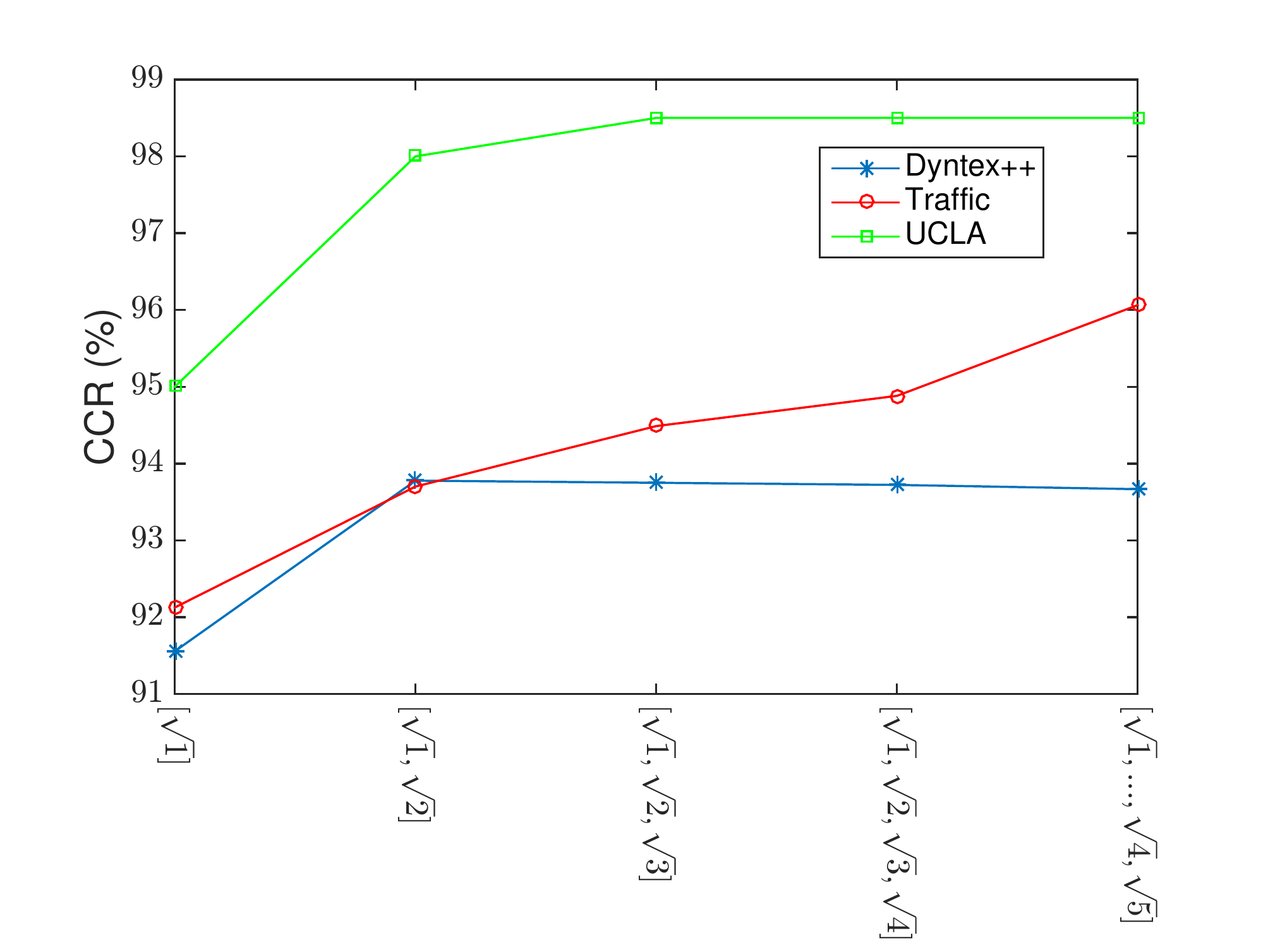}}\\
		\caption{Plots with the parameter analysis in function of the correct classification rate for the Dyntex++, Traffic and UCLA-50 databases.}
		\label{fig:parametros_atividade}
	\end{figure}

	The parameters $\tau_0$, $\tau_i$ and $\tau_f$ indicate the set of thresholds used in the function $\phi(\tau,E)$. These parameters are responsible by a multi-scale analysis of the method. Figure \ref{fig:parametros_atividade} (a) shows the correct classification rate (CCR) according to the initial threshold $\tau_0$ for the three databases. The initial threshold values that obtained the highest CCR were $\tau_0 = 2, 4, 8$, for Dyntex++, Traffic and UCLA-50 databases, respectively. These values show that low initial threshold $\tau_0$ are important because they provide small-scale details in the recognition stage \cite{Goncalves2015211}.

	The incremental threshold $\tau_i$ is responsible for the step in which the thresholds starting from $\tau_0$ to $\tau_f$ will be increased. 
	Thus, the proposed method can provide the analysis of a range of scales. 
	The CCR's for different values of $\tau_i$ in the three databases are plotted in Figure \ref{fig:parametros_atividade} (b). 
	For the Dyntex++, Traffic and UCLA-50 databases, the incremental thresholds values that provided the best CCR were $\tau_i = 48, 110, 96$, respectively. 
	Notice that for the three databases, the best success rate was obtained by high values of $\tau_i$. 
	For high values of $\tau_i$, the method explores various scales, analyzing the variation of patterns of micro to macro details.
	
    To evaluate the final threshold $\tau_f$, it is considered the number of thresholds $n_f$, such that $\tau_f = \tau_0 + n_f *\tau_i$. 
    Thus $n_f$ is the number of thresholds used in the multi-scale analysis. 
    Figure \ref{fig:parametros_atividade} (c) shows the CCR according to $n_f$. 
    The best CCR was achieved for $n_f = 4,1,2$ on the Dyntex++, Traffic and UCLA-50 databases, respectively. 
    Note that the values of $n_f$ obtained were low, in fact, they are associated with high values of the incremental threshold $\tau_i$. 
    Therefore, few scales already provided a good characterization of the dynamic texture, a relevant result with respect to the computational time.
	
	Finally, it is evaluated the radius $r$. The Figure \ref{fig:parametros_atividade} (d), shows the result of different combinations of radiuses.  
    In the Dyntex++ database, the best result was obtained with the concatenation of radiuses $r = [\sqrt{1},\sqrt{2}]$. 
    For the Traffic and UCLA-50 databases, the concatenation that provided the maximum CCR were $r =[\sqrt{1},\sqrt{2},\sqrt{3},\sqrt{4},\sqrt{5}]$ and $r = [\sqrt{1},\sqrt{2},\sqrt{3}]$, respectively.
    
    From the analysis above, it is possible to determine the set of thresholds and radiuses for each database. 
    In general, the number of scales necessary to obtain the maximum CCR in each database is closed. 
    Note that good results are obtained with combinations of thresholds that start with low values, are incremented by high values (average 85) and final values are close to the maximum weight (255). 
    In relation to the radiuses, the combination $r = [\sqrt{1},\sqrt{2},\sqrt{3}]$ provides a good CCR for all databases.

\begin{figure}[!htbp]
		\centering
		\subfigure[Dyntex++ database]{\includegraphics[width=0.9\textwidth]{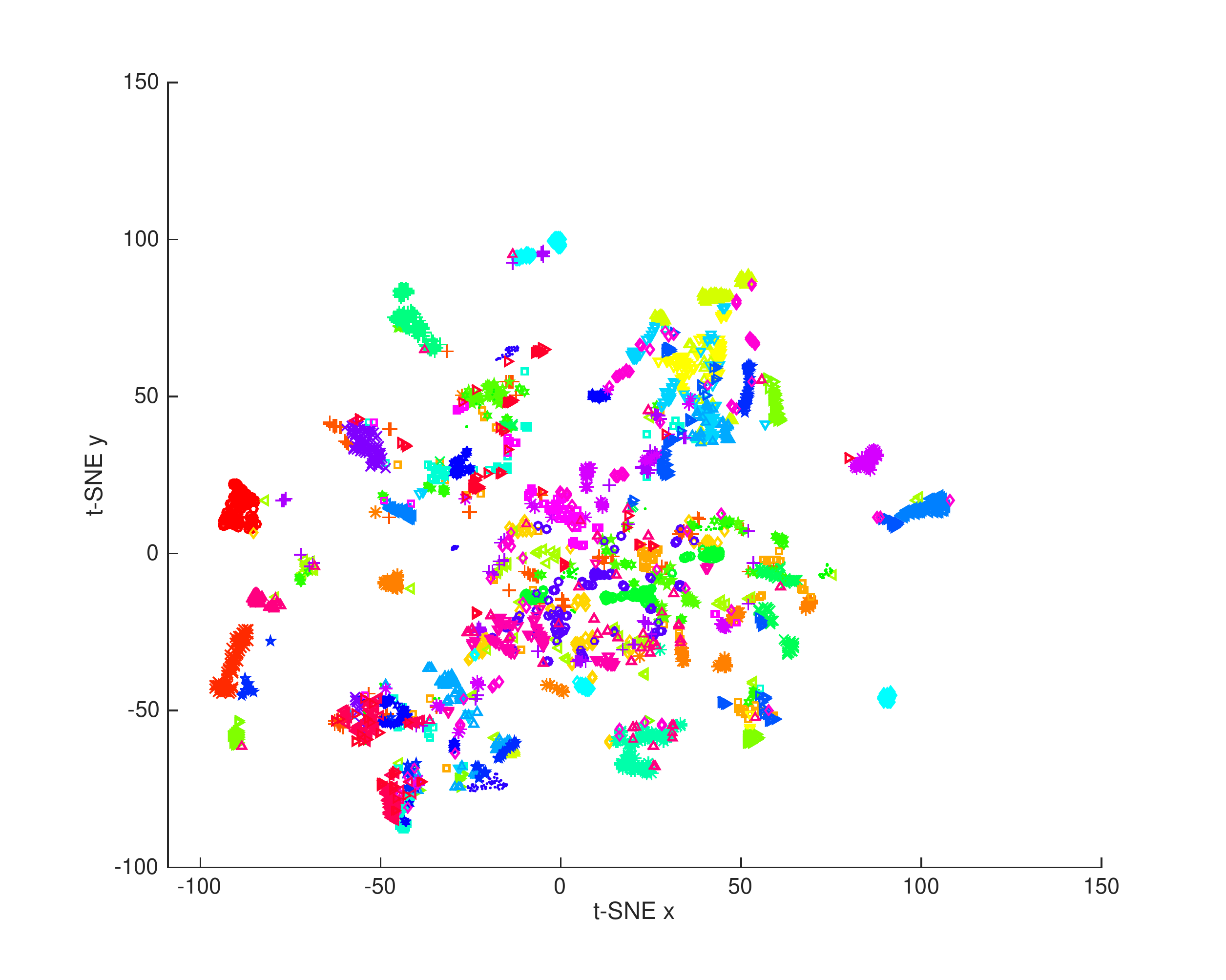}}\\
		\subfigure[UCLA-50 database]{\includegraphics[width=0.49\textwidth]{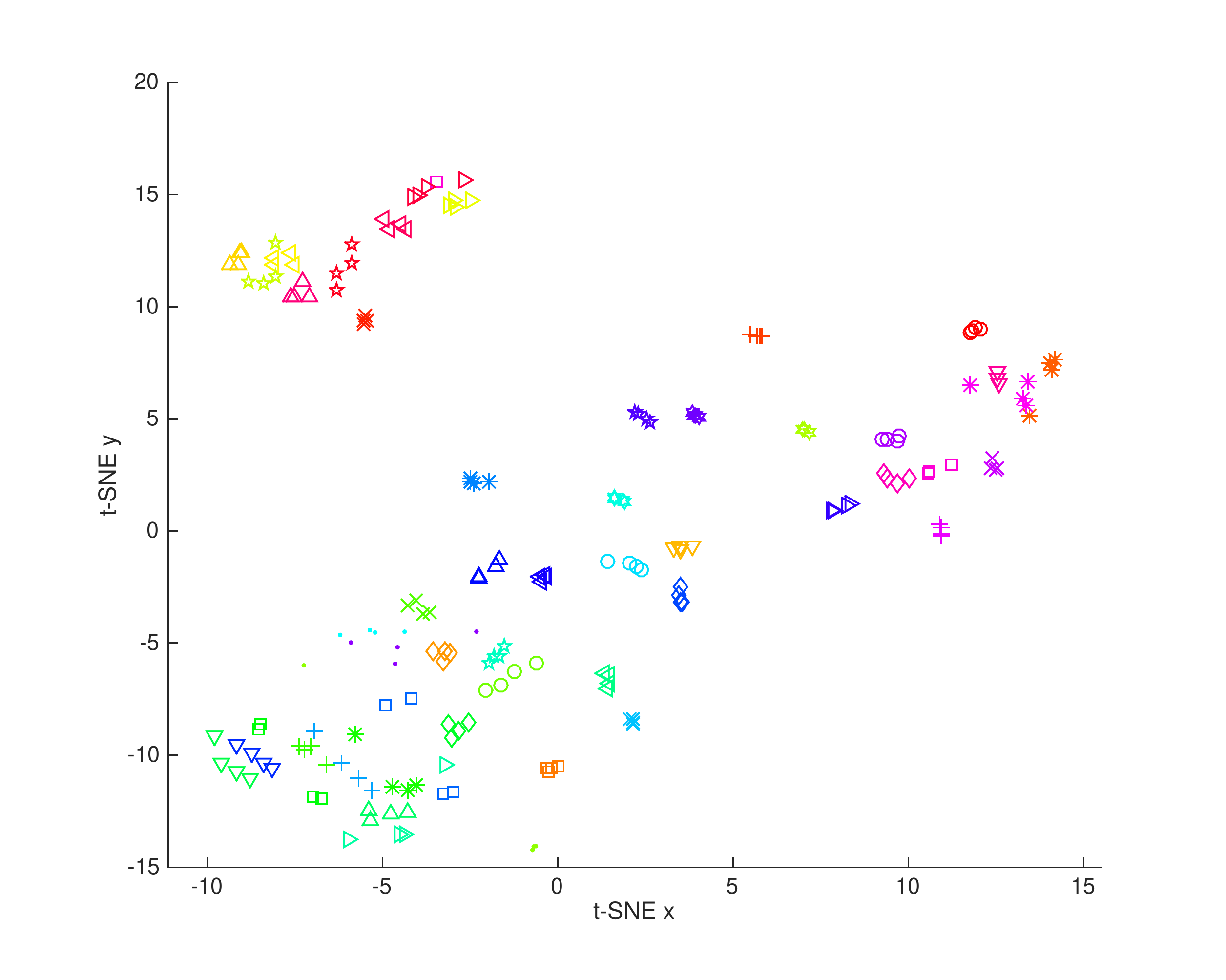}}
		\subfigure[Traffic database]{\includegraphics[width=0.49\textwidth]{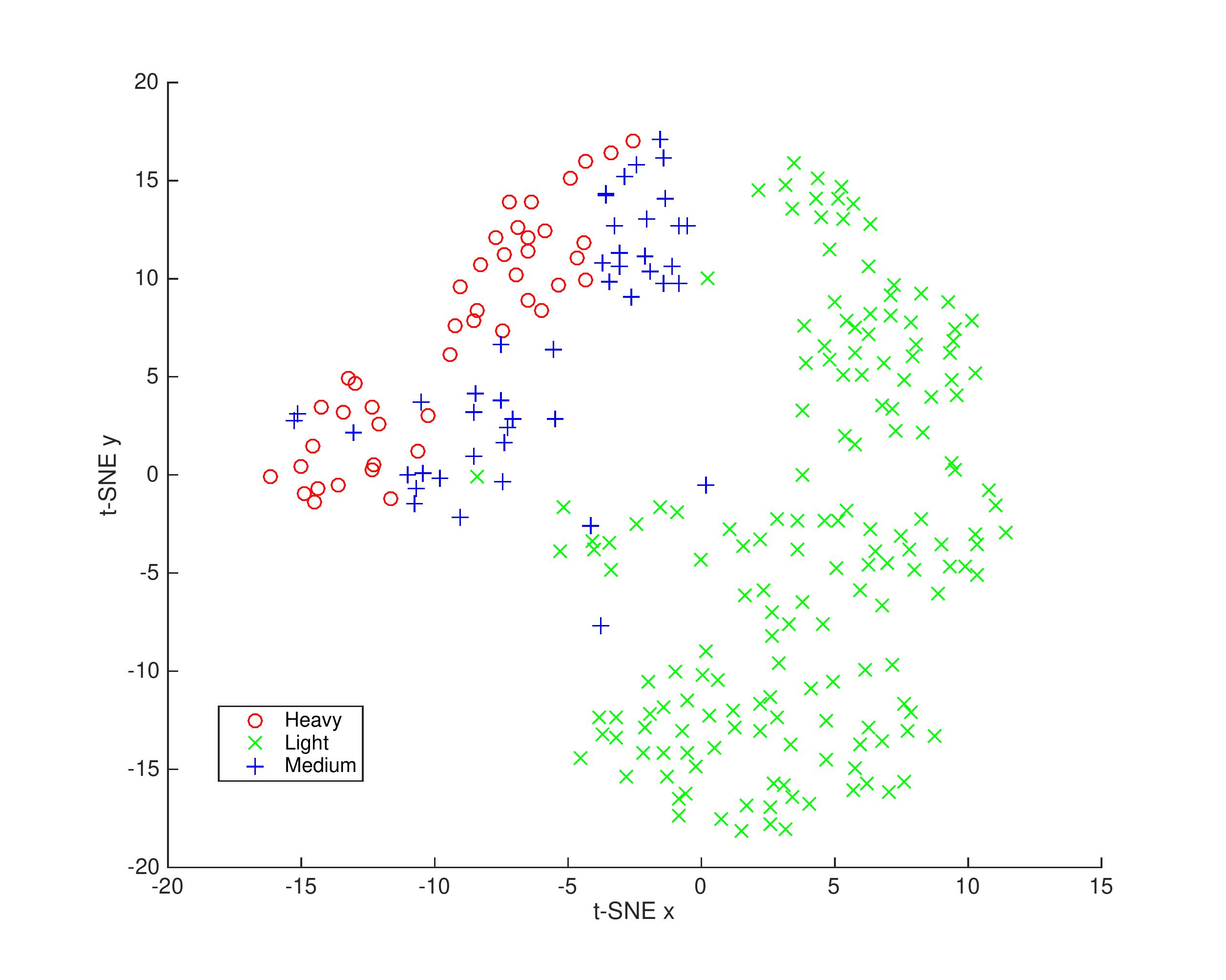}}
		\caption{t-SNE projection of the feature vectors extracted by the proposed approach in the three databases. The feature vectors were extracted using the parameters setup which obtained the best correct classification rate.}
		\label{fig:plot_tsne}    
	\end{figure}

\begin{figure}[!htbp]
		\centering
		\includegraphics[width=1\columnwidth]{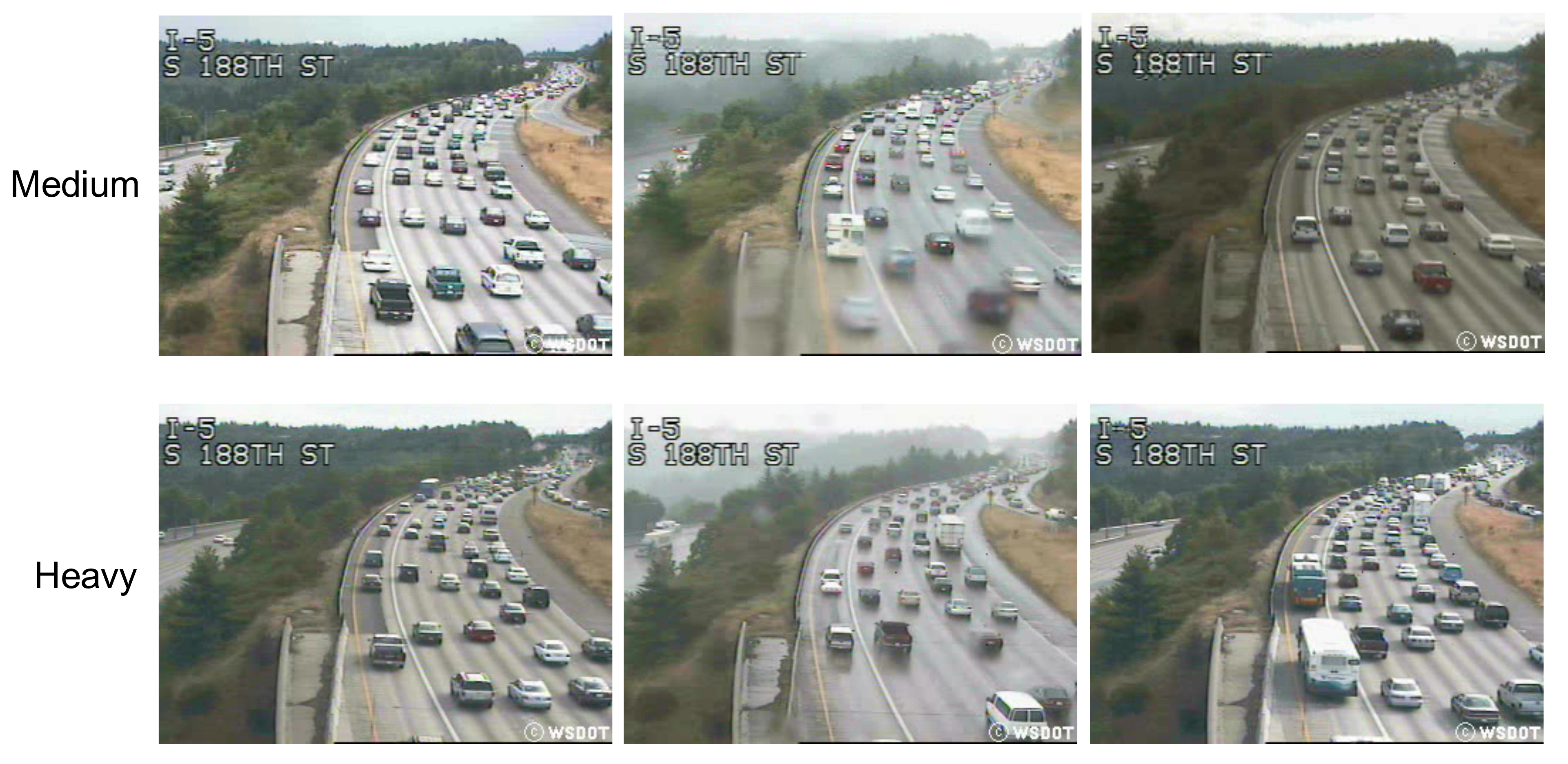}    
		\caption{Examples of the first frame of videos of the medium and heavy traffic classes with similar appearance.}    
		\label{fig:similar_traffic}
	\end{figure}

	The feature vectors built from each sample are embedded in a two-dimensional space for visualization; this is achieved by applying the t-Distributed Stochastic Neighbor Embedding - t-SNE \cite{t-SNE}. 
	It is worth noting that t-SNE uses information of class for display only. 
	The results of the proposed method using features extracted from Dyntex++ and UCLA-50 databases are illustrated in Figures \ref{fig:plot_tsne} (a) and (b), respectively. 
    Figure \ref{fig:plot_tsne} (c) shows the projected feature vectors for Traffic database. These results indicate that medium and heavy traffic classes have characteristics very similar in opposition to the light traffic. Figure \ref{fig:similar_traffic} shows examples of the first frame of some videos from the medium and heavy traffic classes.
	As can be seen, the appearance of videos of the medium traffic class is similar to the heavy traffic, which justifies the proximity of the points in the projection.
	
	\subsection{Rotation invariance}
\begin{figure}[!htbp]
		\centering
		\subfigure[Activity histogram $h^{r,\tau}_s$ ]{\includegraphics[width=0.49\textwidth]{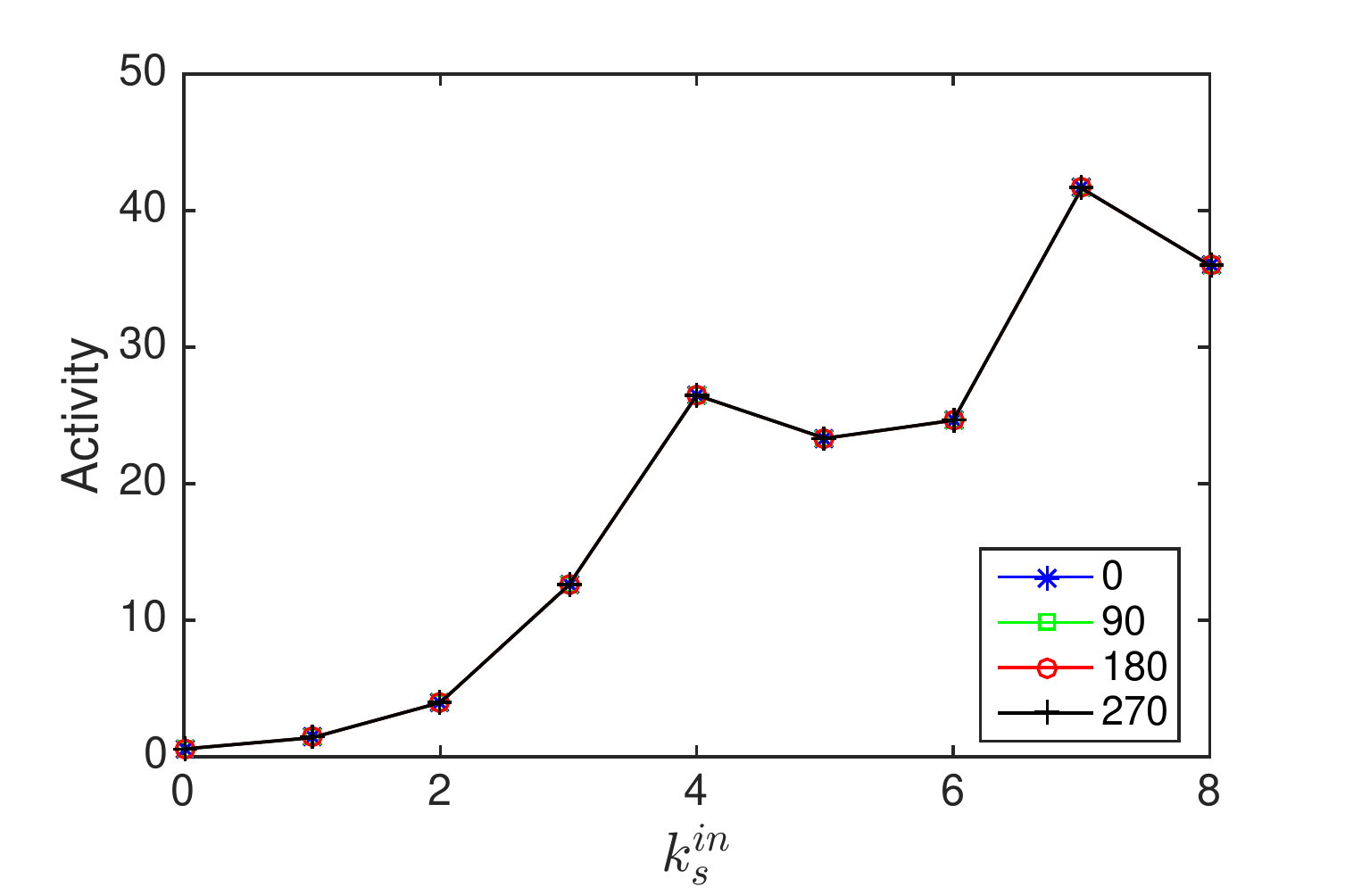}}
		\subfigure[Activity histogram $h^{r,\tau}_t$]{\includegraphics[width=0.49\textwidth]{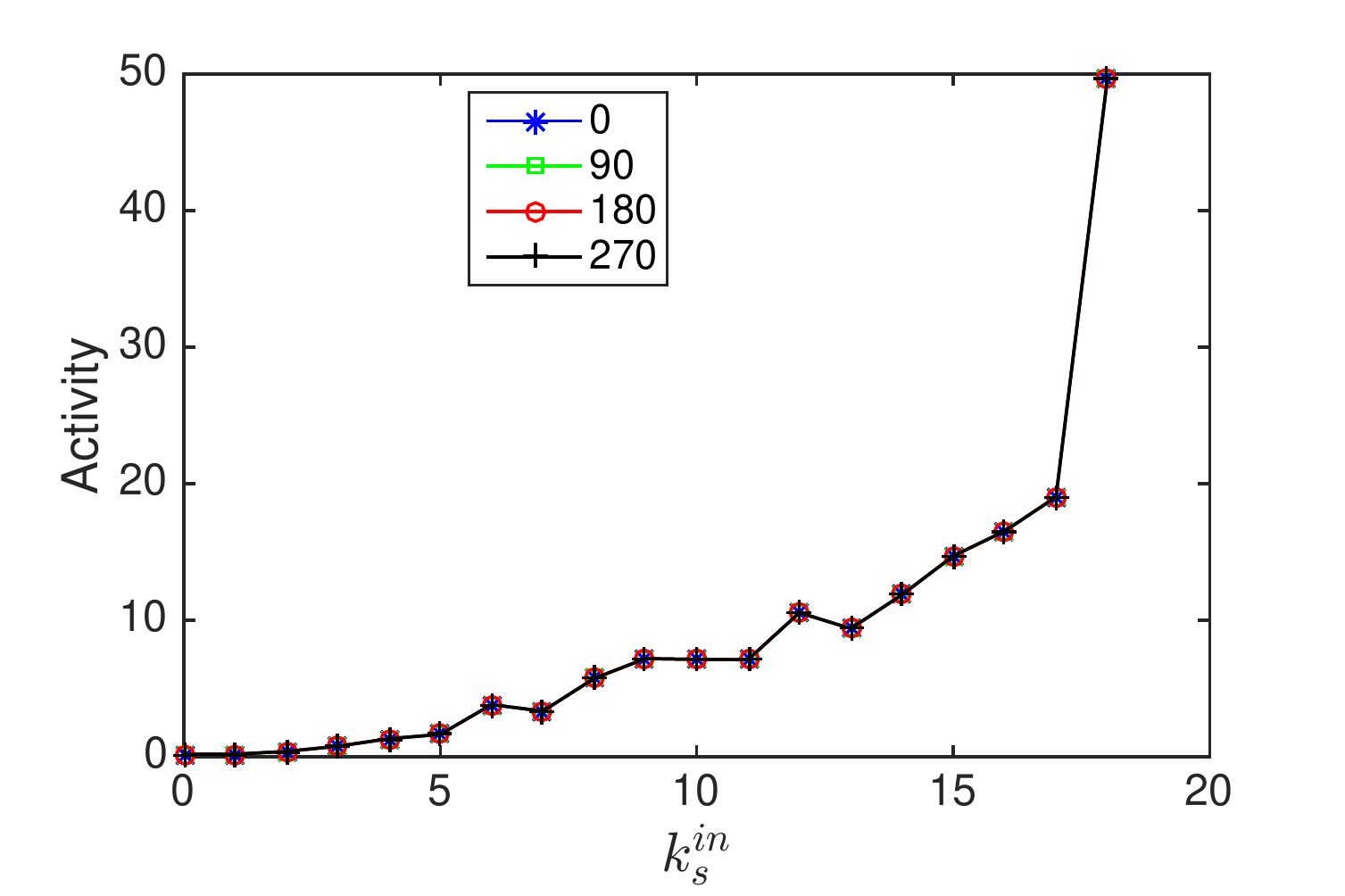}}
		\caption{Histograms $h^{r,\tau}_s$ and $h^{r,\tau}_t$ extracted from dynamic textures rotated.}
		\label{fig:plot_rotate}    
	\end{figure}

	Here, we also present an analysis concerning the rotation invariance of the proposed method, which is an interesting and desirable characteristic for dynamic texture methods.
	Consider a video $V$ and a transformed video $V'$ by a map $M$, such that $V(p)=V(Mp)$. 
	Given two networks $N$ and $N'$ obtained from videos $V$ and $V'$, they must be exactly the same in order to the proposed method be invariant to rotation. To corroborate the invariance to rotation, we show the histograms $h^{r,\tau}_s$ and $h^{r,\tau}_t$ obtained from a video in four rotation. As can be seen in Figure \ref{fig:plot_rotate}, the histograms for the four angles are nearly the same.
	This corroborates the analysis of rotation in networks and diffusion performed in the work of static textures \cite{Backes2013168,Goncalves201651}.
	
	\subsection{Motion analysis}
	
\begin{figure}[!htbp]
		\centering
		\subfigure[Random Motion]{\includegraphics[width=1\textwidth]{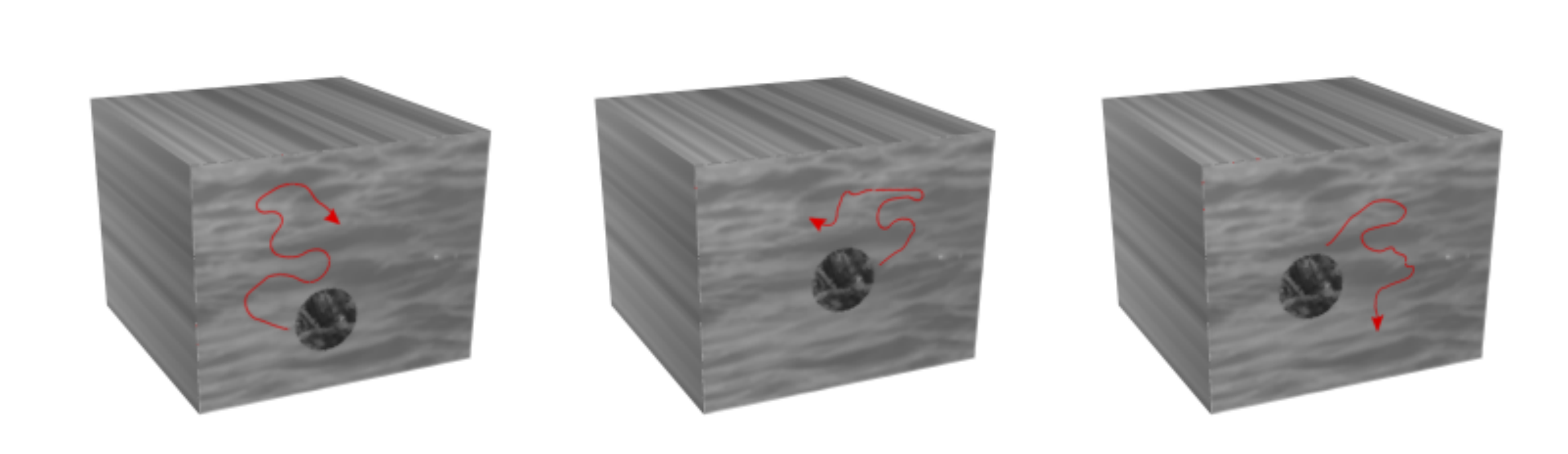}}\\
		\subfigure[Circular Motion]{\includegraphics[width=1\textwidth]{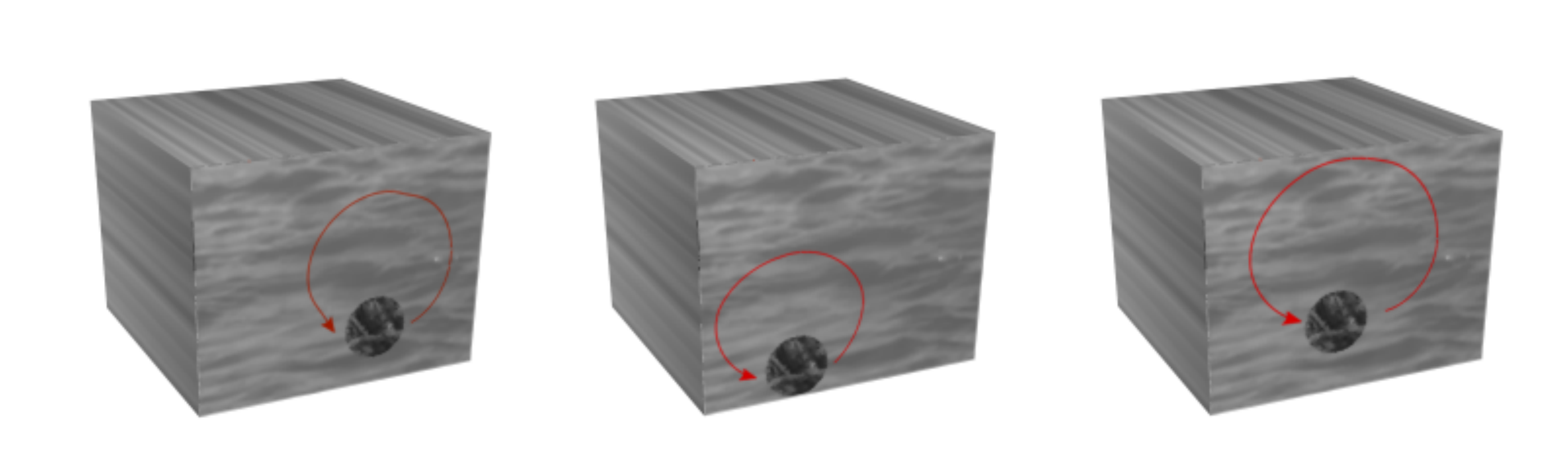}}\\
		\subfigure[Linear Motion]{\includegraphics[width=1\textwidth]{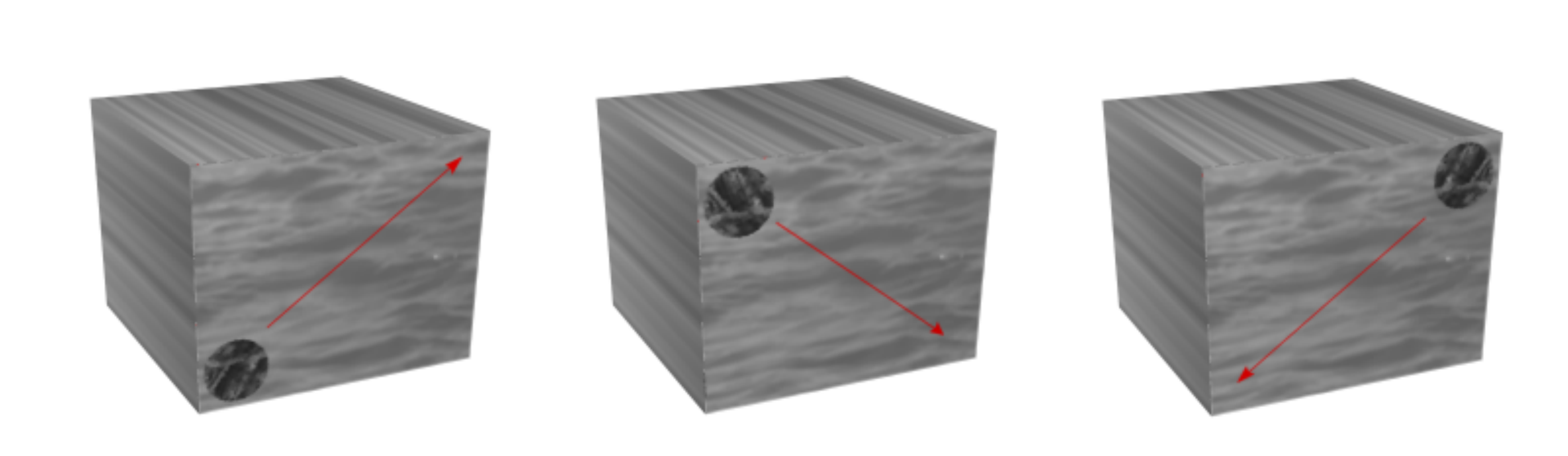}}\\
		\subfigure[Without Motion]{\includegraphics[width=1\textwidth]{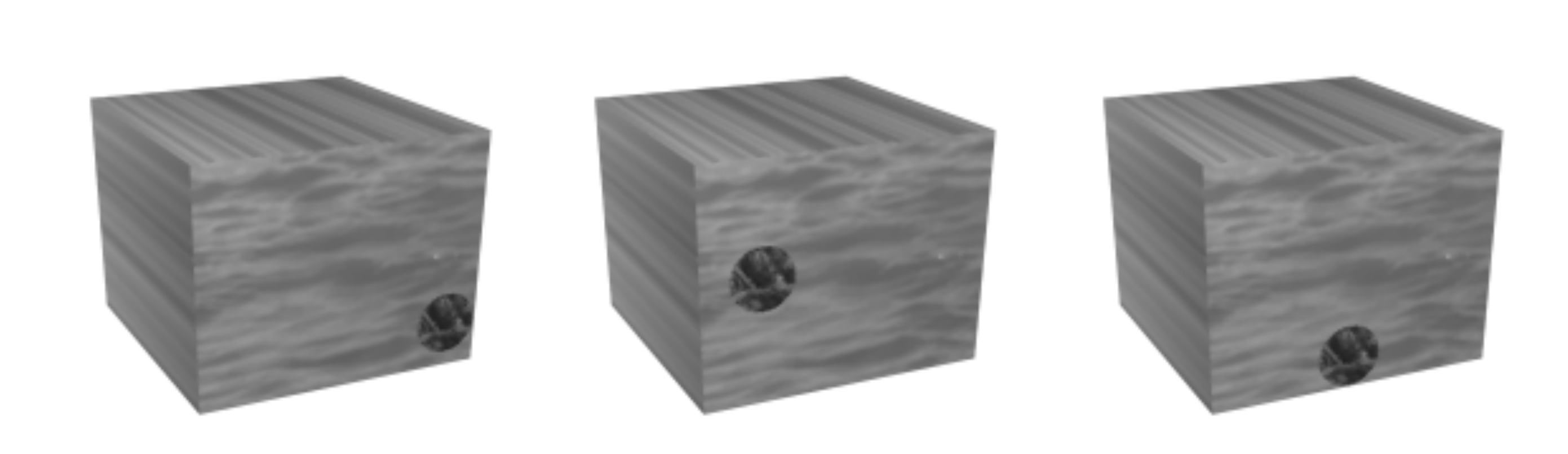}}\\
		\caption{Four classes of synthetic dynamic textures used to evaluate motion features extracted by the proposed approach are shown. Each class contains 4 samples with different motion patterns.}
		\label{fig:text_sintetica}
	\end{figure}

	An important property of dynamic texture methods is their capacity to deal with different motion patterns. 
	In this section, we evaluate the influence of different motion patterns in the features extracted by the proposed method. In the experiments, we have used 120 synthetic dynamic textures of different motion patterns. 
	The synthetic dynamic textures are divided into 4 classes of motion: circular, linear, random and without motion. Each video has approximately 60 frames with a resolution of 180 $\times$ 220 pixels. 
	The video is composed of two classes of dynamic textures, one class is used as background (ocean waves) and the other one (seaweed) is used to simulate motion patterns.
	Figure \ref{fig:text_sintetica} shows examples of synthetic dynamic textures with the motion patterns labeled as red.

	The experiments were conducted as follows. First, the proposed method was applied in each sample of synthetic dynamic texture using the best parameters achieved in the UCLA-50 database ($t_0=8$, $t_i=96$, $n_f=2$ and $r = [\sqrt{1},\sqrt{2},\sqrt{3}]$).  
    We choose this database due to the different motion patterns present in it and the good results achieved by the proposed method.
    Since the goal is to analyze motion patterns, we considered only the feature vector $h^{r,\tau}_t$ for dynamic texture representation.  For purposes of visualization, the features vector $h^{r,\tau}_t$ are embedded in a two-dimensional space through the application of the t-Distributed Stochastic Neighbor Embedding - t-SNE \cite{t-SNE}. The t-SNE is a technique for visualization of high-dimensional data giving each data point a location on a two-dimensional map. It is important to mention that the t-SNE does not use information of class to perform the projection. Therefore, the samples are divided into classes only for visualization. 

\begin{figure}[!htbp]
		\centering
		\includegraphics[scale=0.5]{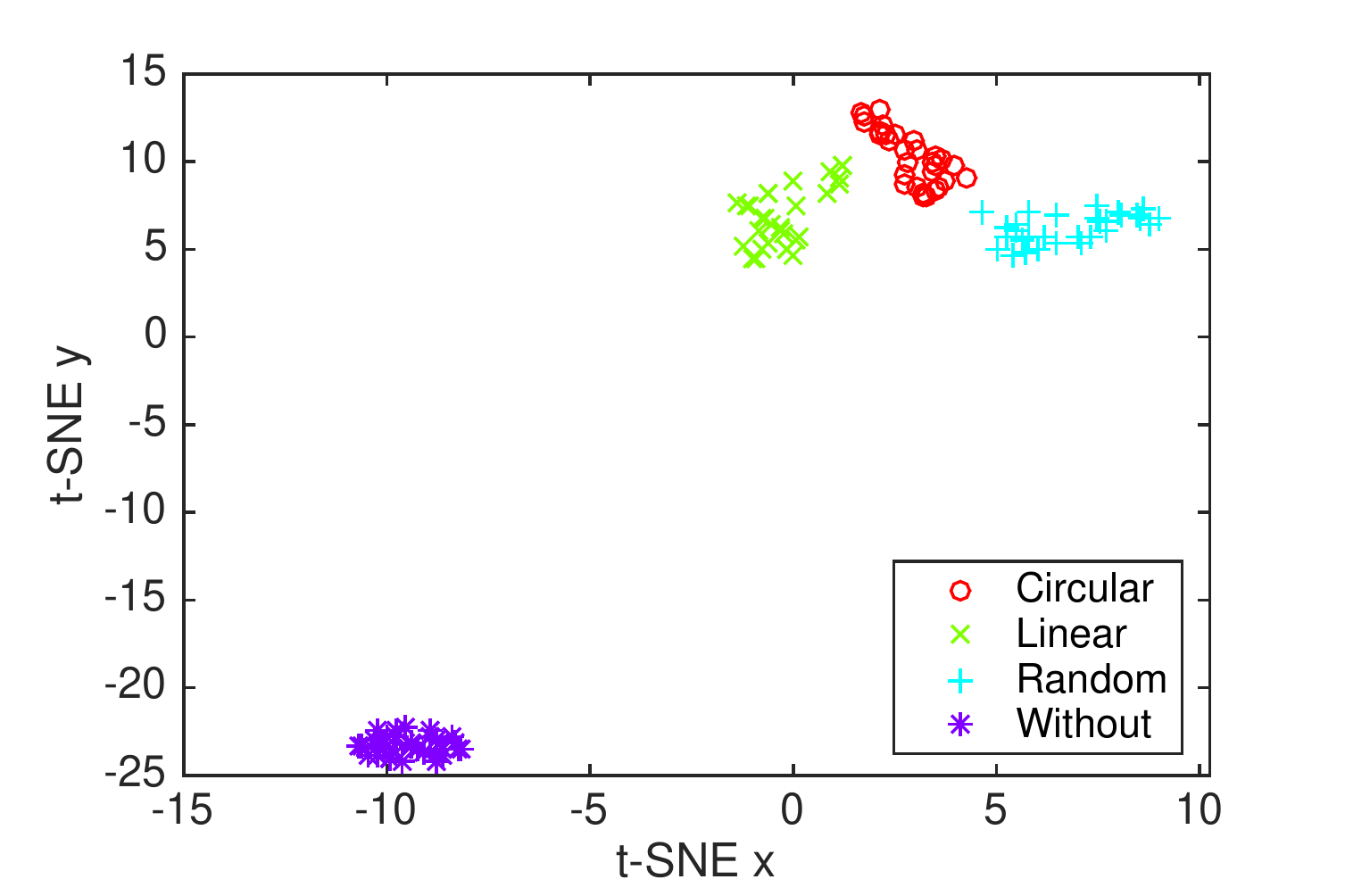}
		\caption{t-SNE projection for features extracted from different classes of motion.}
		\label{fig:pca_movimento}
	\end{figure}

Figure \ref{fig:pca_movimento} shows the projection of the features extracted by the proposed method. Note that, as expected the samples with the same motion pattern are clustered in the projected space.  In this way, it is clear that the proposed method can separate the four classes for which the main characteristic rely on motion patterns. Another desirable property in the methods of dynamic texture characterization is the ability to deal with video transformations such as rotation. As we can see in Figure \ref{fig:text_sintetica} the linear motion is rotated. However, the projection indicates that the features remain clustered. It occurs because the agent is not influenced by rotated pixels since it searches for the next vertex in all directions and based on a random raffle and a probability.
	
	\subsection{Comparison with other methods}

	In this section, we use the parameters obtained by the proposed method in the previous sections and we compare its results with the literature methods. In order to perform the comparison, it was considered the following measures: number of features and correct classification rate (CCR).
 The results of the VLBP, LBP-TOP, CDT-TOP and CNDT methods were obtained from our own implementation. On the other hand, the results of remaining methods are from the original literature. All results were obtained using the same experimental setup.

\begin{figure}[!htbp]
		\centering
		\includegraphics[width=.9\columnwidth]{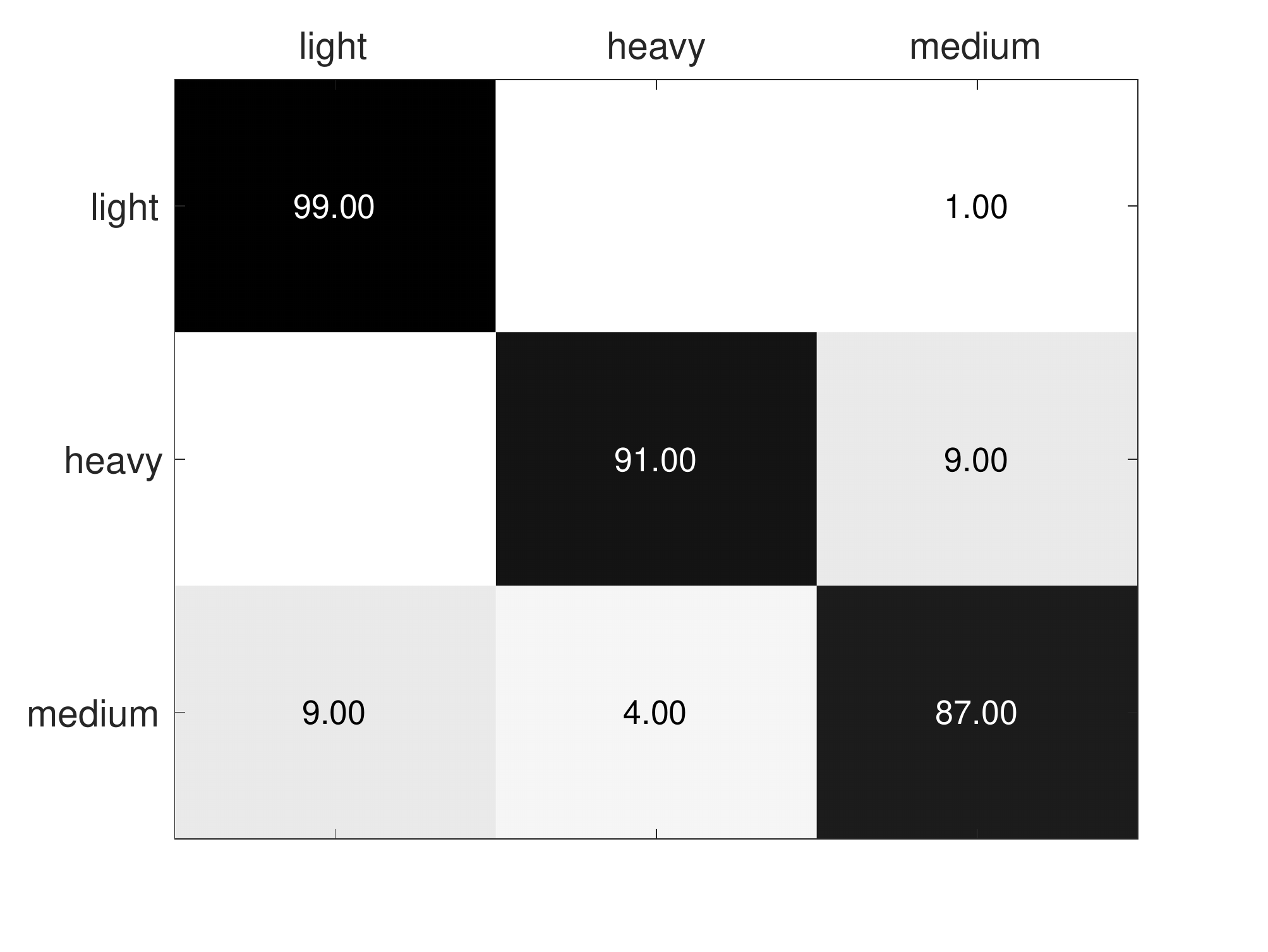}    
		\caption{Confusion matrix (\%) for traffic database, 96.06\%.}    
		\label{fig:conf_traffic}
	\end{figure}

    Tables \ref{tab:comp_traffic} shows the performance of the methods for the Traffic database. 
    In this database, the proposed method and CNDT method has relatively better performance than the other methods. 
    Although the size of the feature vector of the CNDT method is the double of the proposed method, our method is approximately three times faster to feature extraction, as can be seen in Section \ref{sec:time}.  
    In relation to the parameters, the two methods have four parameters to build the network.
    However, in contrast to the CNDT method,  the proposed method models the video in a directed network, consuming the half of memory space.
    Figure \ref{fig:conf_traffic} shows the confusion matrix of the Traffic database. 
    Note that the medium traffic class is most difficult to describe and classify, which can also be seen from Figure \ref{fig:plot_tsne} (c). Noticed that the two classes, medium and heavy traffic present high similarity (see Figure \ref{fig:similar_traffic}) and, in this way, can be incorrectly classified.

\begin{figure}[!htbp]
		\centering
		\includegraphics[width=1\columnwidth]{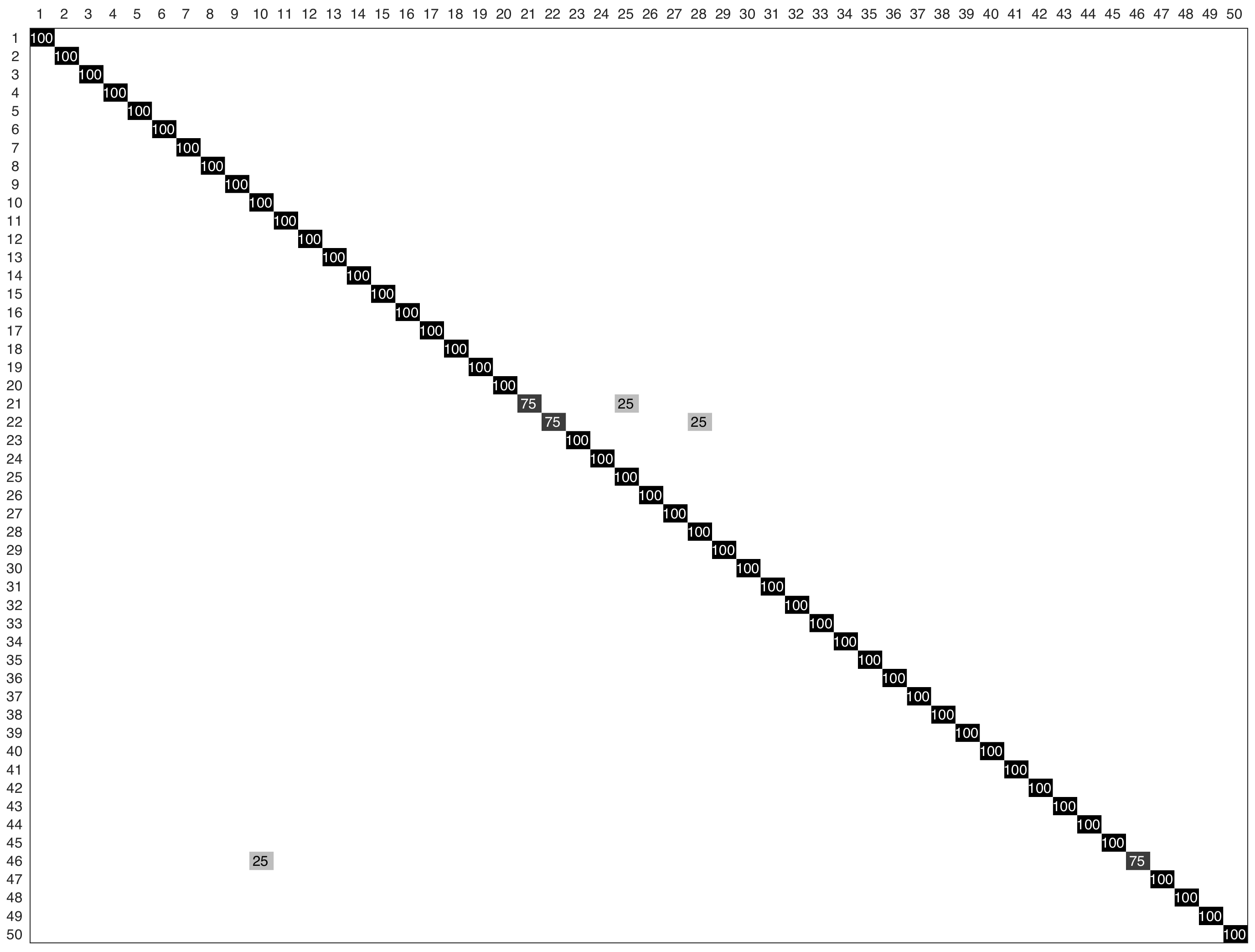}    
		\caption{Confusion matrix (\%) for UCLA database of best CCR, 98.5\%.}
		\label{fig:conf_ucla}    
	\end{figure}

    A comparison of the proposed method against other methods on the UCLA-50 database is shown in Table \ref{tab:comp_ucla50}.
    It is evident that the proposed method obtains a significant improvement compared to the other methods. Also, the sizes of feature vectors of the proposed and CDT-TOP method are considerably lower than other methods.
    The confusion matrix of the UCLA database is shown in Figure \ref{fig:conf_ucla}. 
    As can be noticed, only three samples were incorrectly classified.
       
    Table \ref{tab:comp_ucla9} presents the performance of the proposed and compared methods on the UCLA-9 database.
    As can be seen, the proposed method outperforms all compared methods.
    Besides that, our method and the CDT-TOP method have a smaller number of feature when compared to other approaches.
    It is important to emphasize that the dimension of the feature vector is very important in real-time applications where the time is crucial.
    A comparison of the proposed method and others for the UCLA-8 database is shown in Table \ref{tab:comp_ucla8}. 
    Once again, the proposed method outperform the other methods with approximately 0.42\% (3D-OTF) and 0.57\% (CVLBP) margins.
    It can also be observed that the proposed method performs better than the other LBP based approaches.

\begin{figure}[!htbp]
		\centering
		\includegraphics[width=1\columnwidth]{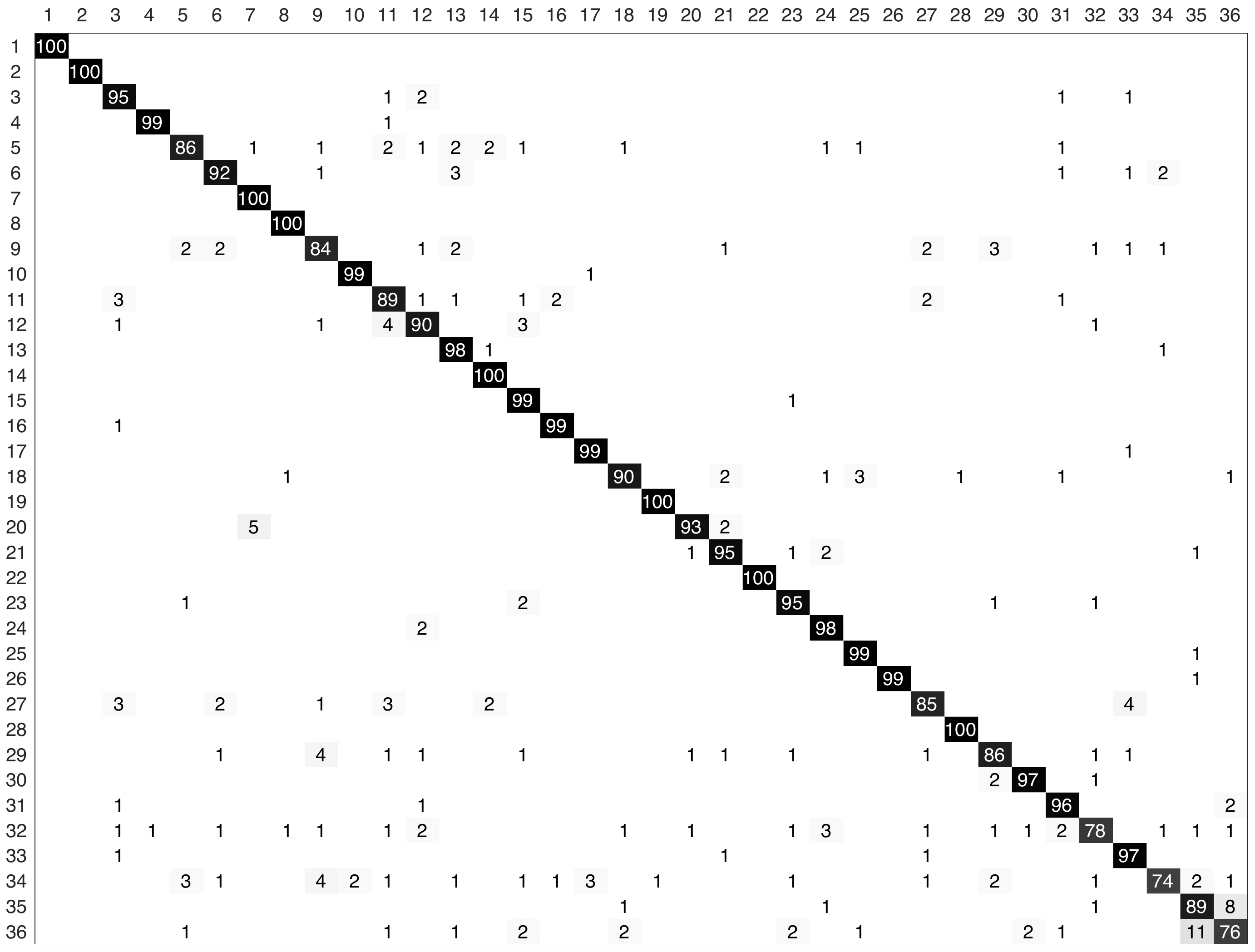}    
		\caption{Confusion matrix (\%) for Dyntex++ database, 93.80\%.}    
		\label{fig:conf_dyntex}
	\end{figure}

The comparison results on the Dyntex++ database is shown in Table \ref{tab:comp_dyntex}.
As observed from the table, the proposed method outperforms the CDT-TOP and CNDT methods. 
The VLBP and LBP-TOP methods achieved the higher correct classification rate, 96.14\% and 97.72\%, respectively.
However, the proposed method extracts 141 features, while the size of the feature vector of the RI-VLBP and LBP-TOP methods are 16384 and 768, respectively. 
High dimensional feature vectors are computationally expensive, increasing the time required to classify a DT and memory consumption.  
Once dynamic textures have been used in several real-time expert systems, high dimensional feature vectors can be impractical.
For instance, in fire detection system is very important the computational time to trigger the alarm in time. 
In addition, the proposed method is approximately twice faster than the VLBP method.
In terms of parameters, as the proposed method the VLBP and LBP-TOP methods have parameters.
The VLBP method has three parameters that can imply in the size of feature vector and complexity considerably, while the LBP-TOP method has two parameters that can also vary in each orthogonal planes. 
These may influence the computational and accuracy performance of the methods.
Figure \ref{fig:conf_dyntex} shows the confusion matrix of the Dyntex++ database where the samples incorrectly classified are distributed to several classes. 
Indeed, Dyntex++ is the most difficult database to be classified.
	
	\begin{table}[!htbp]
		\centering
		\caption{Classiication results of the methods on Traffic database.}
		\begin{tabular}{lll} \hline
			Methods      & N. of Features & CCR (\%)  \\ \hline
			RI-VLBP \cite{Zhao2007}        & 16,384     & 93.31 ($\pm$ 4.34)       \\
			LBP-TOP \cite{ZhaoLPB07}  & 768       &  93.70 ($\pm$ 4.70)     \\  
			CDT-TOP \cite{Goncalves2013turistaESWA}  & 75       & 93.70 ($\pm$ 4.83)      \\  
			CNDT \cite{Goncalves2015211} & 144       & 96.46 ($\pm$ 4.10)   \\      
			Proposed method     & 297      & 96.60 ($\pm$ 4.38)   \\ \hline
		\end{tabular}
		\label{tab:comp_traffic}
	\end{table}
    
    \begin{table}[!h]
	\centering
	\caption{Results of the methods on the UCLA-50 database using 4-fold cross validation.}
	\label{tab:comp_ucla50}
	\begin{tabular}{lll} \hline
		Method      & Number of descriptors &  CCR (\%)  \\ \hline
        KDT-MD \cite{chan2007classifying} & - & 89.50 \\
        DFS \cite{dyntexfractal} & - & 89.50 \\
        3D-OTF \cite{3DOTF} & 290 & 87.10 \\
        CVLBP \cite{Tiwari2016CVLBP} & -& 93.00\\
        RI-VLBP \cite{Zhao2007}  & 16384   &   77.50 ($\pm$ 8.98)       \\
		LBP-TOP \cite{ZhaoLPB07}  & 768       &   95.00 ($\pm$ 4.44)   \\  
		CDT-TOP \cite{Goncalves2013turistaESWA}  & 75       &  95.00 ($\pm$ 4.78)      \\  
		CNDT \cite{Goncalves2015211} &  420       &  95.00 ($\pm$ 5.19)           \\
        Proposed method & 169 & 98.50 ($\pm$ 3.37) \\
        \hline
	\end{tabular}
    \label{tab:ucla50}
\end{table}

\begin{table}[!h]
	\centering
	\caption{Comparison of the proposed method with other dynamic texture methods on the UCLA-9 database. }
	\label{tab:comp_ucla9}
	\begin{tabular}{lll} \hline
		Method      & Number of descriptors & CCR (\%)  \\ \hline
        3D-OTF \cite{3DOTF} & 290 & 96.32 \\
        CVLBP \cite{Tiwari2016CVLBP} & - & 96.90\\
        High level feature \cite{wang2015exploiting} & - & 92.60 \\
		WMFS \cite{wmfsji2013wavelet} & 702
        & 96.95 \\
        Chaotic vector \cite{wang2016chaotic} & 300 & 85.10 \\
   		RI-VLBP \cite{Zhao2007}        & 16384     &   96.30        \\
		LBP-TOP \cite{ZhaoLPB07} & 768       &   96.00    \\       
		CDT-TOP \cite{Goncalves2013turistaESWA} & 75       &   96.33 ($\pm$ 2.46)      \\  
		CNDT \cite{Goncalves2015211} &  336       &  95.61 ($\pm$ 2.72)     \\ 
                Proposed method & 169 &  97.80 ($\pm$ 1.53)\\
        \hline
      
	\end{tabular}
    \label{tab:ucla9}
\end{table}

\begin{table}[!h]
	\centering
	\caption{Comparison results for existing methods on the UCLA-8 database.}
	\label{tab:comp_ucla8}
	\begin{tabular}{lll} \hline
		Method      & Number of descriptors & CCR (\%)  \\ \hline
        3D-OTF \cite{3DOTF} & 290 & 95.80 \\
        CVLBP \cite{Tiwari2016CVLBP} & - & 95.65\\
        High level feature \cite{wang2015exploiting} & - & 85.65\\
        Chaotic vector \cite{wang2016chaotic} & 300 & 85.00 \\
   		RI-VLBP \cite{Zhao2007}        & 16384     &   91.96     \\
		LBP-TOP \cite{ZhaoLPB07} & 768       &   93.67    \\       
		CDT-TOP \cite{Goncalves2013turistaESWA} & 75       &  93.41 ($\pm$ 6.01)      \\  
		CNDT \cite{Goncalves2015211} &  336       &  94.32 ($\pm$ 4.18)         \\
         Proposed method & 169 &  96.22 ($\pm$ 4.80)\\

        \hline
	\end{tabular}
    \label{tab:ucla8}
\end{table}

    \begin{table}[]
	\centering
	\caption{Experimental results for dynamic texture methods on the Dyntex++ database.}
	\label{tab:comp_dyntex}
	\begin{tabular}{lll} \hline
		Method      & Number of descriptors & CCR (\%)  \\ \hline
		RI-VLBP \cite{Zhao2007}       & 16384     &  96.14 ($\pm$ 0.77)        \\
		LBP-TOP \cite{ZhaoLPB07} & 768       &  97.72 ($\pm$ 0.43)      \\  
        CDT-TOP \cite{Goncalves2013turistaESWA} & 75       & 91.39 ($\pm$ 1.29)       \\  
		CNDT \cite{Goncalves2015DyntexNeu} &  336      &  83.86 ($\pm$ 1.40)      \\ 
                Proposed method &  141 & 93.80 ($\pm$ 1.08)\\
        \hline
     
	\end{tabular}
    \label{tab:dyntex++}
\end{table}

	\subsection{Computational complexity and processing time} \label{sec:time}
	
	Basically, the proposed method models a dynamic texture with $n \times n \times T$ pixels in a directed network. 
	Next, $M$ random walks of size $l$ are initiated in each vertex to estimate its activity. 
	Each pixel is mapped to a vertex, which is connected with the  $(2r+1)\times (2r+1)\times (2r+1)$ neighbors. 
	Therefore, to build the network $O((2r+1)\times (2r+1)\times (2r+1) \times n^2 \times T )$ operations are required. 
	However, in practice, the radius $r$ is much smaller than the number of pixels in the video, such that $r\ll n^2 \times T$.
	
	The computational complexity of the proposed method is given then by $O(M \times l \times n^2 \times T)$. 
	Since $M$ is a multiplicative constant (in this work $M=50$) and $M \ll n^2\times T $, it can be disregarded.
	The other variable that affects the complexity is the length of the walk $l$. The best case is when the vertices are disconnected, and in this case, the length of the walk is $l=1$, leading to complexity equal to $O(n^2 \times T)$. 
	The worst case occurs when the length of all walks is the maximum, that is equal to $L$ (this work $L = 1000$). 
	The complexity of the worst case is $O(L \times n^2 \times T )$.

\begin{figure}[!htbp]
		\centering
		\includegraphics[width=.7\columnwidth]{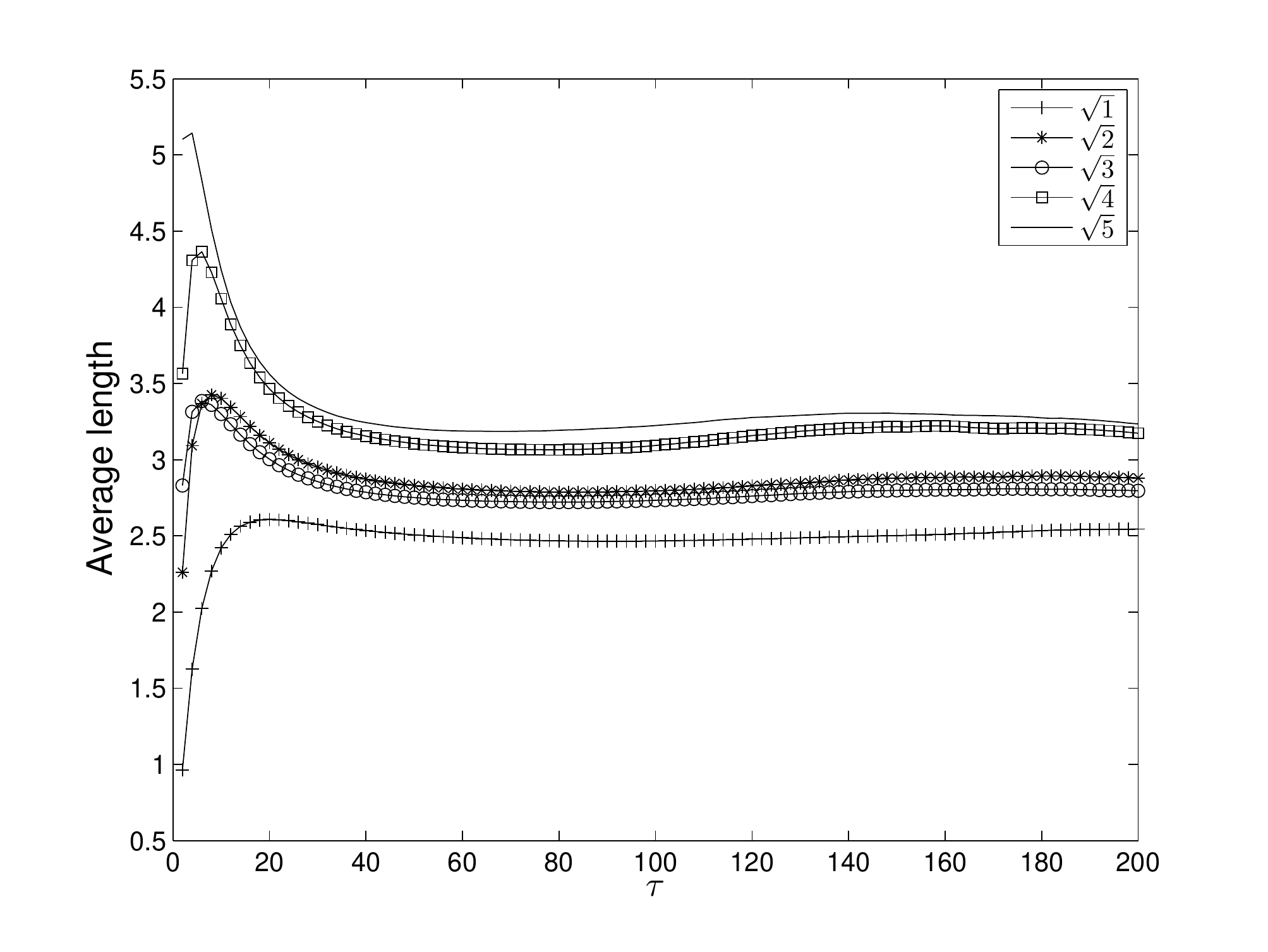}
		\caption{Average walk length in the Dyntex++ database for various values of threshold and radius.}
		\label{fig:length_walk}
	\end{figure}

    The average case is evaluated in Dyntex database which has 3600 video samples of $50\times50\times50$ each. 
	Figure \ref{fig:length_walk} shows the average walk length to different radius $r$ and thresholds $\tau$. 
	In general, the average length of the walking is below 5, for instance, using $r=\sqrt{4}$ and $\tau = 20$, the average walk length is $l = 3.46$. 
	This leads to a computational complexity close to the best case, which is $O(n^2 \times T )$. 
 	
 For comparison purposes, the CNDT and CDT-TOP methods also have a complexity of order $O(n^2 \times T )$, considering the average case. 
   The LBP-TOP method and their extensions (CVLBP method) have complexity in order of $O(n^2 \times T \times 2^P)$, where $P$ defines the number of neighboring points.
   On the other hand, the complexity of the VLBP and CVLBP methods are of order $O(n^2 \times T \times 2^{3P})$. In the case of values of $P$ greater than 1 the complexity and size of the feature vector increase considerably. 
   These complexities show that the proposed method is competitive in terms of computational complexity.

  In terms of processing time, to build the complex network with radius $r = \sqrt{5}$ (worst case) of a video with $50 \times 50 \times 50$ vertices, the proposed method took on average 0.368 s using Intel (R) Core (TM) i7-3610QM CPU @ 2.30Ghz, 16 GB RAM, 64-bit Operating System. 
    For feature extraction considering the Dyntex++ database parameters, the proposed method took on average 4.830 s. 
    For comparison purposes, the VLBP, LBP-TOP, CDT and CNDT methods obtained on average 8.04, 0.51, 15.03 and 22.33 seconds, respectively. 
    The high processing time of the CNDT method is due to the large number of radiuses needed to build the network (7 radiuses), increasing the complexity of the method.  
    Although the proposed method has not obtained the lower running time, the results indicate competitive running time for real-time application compared to other methods.

	
	\section{Conclusion}
    
    In this paper, we proposed a new method for characterization of dynamic textures using the complex network theory. 
   The main contribution of this paper is the use of directed networks to dynamic textures modeling, which tends to provide a better performance than undirected networks \cite{Goncalves201651,diffusionCosta}. Another contribution is the use of diffusion in network to dynamic textures characterization, which provides a better characterization than other traditional network measures used in \cite{Goncalves2015211}, improving the recognition task performance. This follows from the fact that directed diffusion has been used to highlight the
 diversity and separation of the dynamics in the respective network\cite{Goncalves201651,diffusionCosta}.
    The dynamic textures are represented in directed complex networks and its activity associated to the spatial and temporal in-degree is used to compose the feature vector. 
    We demonstrated how the diffusion can be effectively used for characterizing and analyzing complex networks derived from videos of dynamic textures.
    The information in different scales obtained by the proposed method demonstrated that it can provide valuable information about the structure being analyzed.
    
    The proposed method achieved the same result of the compared methods on the Traffic database and its performance on the Dyntex++ although was outperformed by two of the compared methods, the difference was small (close to 4\%). However, the size of the feature vector of other literature methods (e.g. RI-VLBP and LBP-TOP) is much higher as compared to the proposed method, which offset its improvement over our method. This makes the proposed method competitive in expert systems that use videos.
    On the other hand, the proposed method outperformed all the other methods in the UCLA-50, UCLA-9 and UCLA-8 databases.
The proposed method also outperformed the CNDT method on the UCLA and Dyntex++ databases and obtained a similar correct classification rate on the Traffic database. The CNDT method uses undirected network and traditional measurements to analysis, therefore, it is demonstrated that the use of directed network and diffusion improves the recognition performance.
   The proposed method also demonstrated to have desirable properties that are missing in many methods of the literature, such as properly analyze appearance and motion characteristics and rotation invariance.    
       
   Considering the experiments of rotation, the motion analysis, and the good performance obtained in three dynamic texture databases, we can conclude that the method is robust and it can be considered a very good option to dynamic texture recognition.
	Complex Networks based methods have demonstrated a very good performance on the characterization of dynamic textures, which turns out into a promising research field. 
  
 Once the proposed modeling need of four parameters, in future, new ways of dynamic texture modeling in network can be investigated to improve the representation of texture patterns and optimize the number of parameters.  A future related work would be to investigate a function of automatic threshold selection, optimizing the parameters and improving the performance.
 Another research issue is to evaluate other strategies to extract information from the activity joint distribution.
  In addition, as part of the future work, we plane to use new methods of pattern recognition to describe the network. Currently, there is a gap of pattern recognition methods in complex networks in literature. We believe that more sophisticated methods can provide a better classification performance.


	\subsection*{Acknowledgments.} Lucas C. Ribas gratefully acknowledges CAPES (Grant Nos. 9254772/m) and S\~ao Paulo Research Foundation (FAPESP) (Grant No. 2016/23763-8) for financial support. 
	Odemir M. Bruno thanks the financial support of CNPq (Grant \# 307797/2014-7) and FAPESP (Grant \# 14/08026-1).
	Wesley N. Gon\c{c}alves acknowledges support from Fundect (Grant No. 071/2015) and CNPq (Grant No. 304173/2016-9).

\bibliographystyle{abbrv}

\end{document}